\documentclass[12pt,a4paper]{article}

\usepackage{fullpage}
\usepackage{authblk}
\usepackage[english]{babel}
\usepackage[utf8x]{inputenc}
\usepackage{amsmath}
\usepackage{amssymb}
\usepackage{graphicx}
\graphicspath{ {images/} }
\usepackage{longtable}
\usepackage{multirow}
\usepackage{hyperref}
\usepackage{subcaption}
\usepackage{mathtools}
\providecommand{\keywords}[1]{\textbf{\textit{Index terms---}} #1}
\DeclarePairedDelimiter\floor{\lfloor}{\rfloor}

\title{MultiResUNet : Rethinking the U-Net Architecture for Multimodal Biomedical Image Segmentation}
\author[1]{Nabil Ibtehaz}
\author[1,*]{M. Sohel Rahman}

\affil[1]{Department of CSE, BUET,\protect\\ ECE Building, West Palasi, Dhaka-1205, Bangladesh}
\affil[*]{Corresponding author}

\affil[ ]{}
\affil[ ]{\href{mailto:1017052037@grad.cse.buet.ac.bd}{\nolinkurl{1017052037@grad.cse.buet.ac.bd}} , \href{mailto:msrahman@cse.buet.ac.bd}{\nolinkurl{msrahman@cse.buet.ac.bd}}}

\begin{document}
\maketitle

\begin{abstract}

In recent years Deep Learning has brought about a breakthrough in Medical Image Segmentation. U-Net is the most prominent deep network in this regard, which has been the most popular architecture in the medical imaging community. Despite outstanding overall performance in segmenting multimodal medical images, from extensive experimentations on challenging datasets, we found out that the classical U-Net architecture seems to be lacking in certain aspects. Therefore, we propose some modifications to improve upon the already state-of-the-art U-Net model. Hence, following the modifications we develop a novel architecture MultiResUNet as the potential successor to the successful U-Net architecture. We have compared our proposed architecture MultiResUNet with the classical U-Net on a vast repertoire of multimodal medical images. Albeit slight improvements in the cases of ideal images, a remarkable gain in performance has been attained for challenging images. We have evaluated our model on five different datasets, each with their own unique challenges, and have obtained a relative improvement in performance of 10.15\%, 5.07\%, 2.63\%, 1.41\%, and 0.62\% respectively.

\end{abstract}

\keywords{Convolutional Neural Networks, Medical Imaging, Semantic Segmentation, U-Net}

\clearpage
\section{Introduction}

Since the introduction of digital medical imaging equipments, significant attention has been drawn towards applying image processing techniques in analyzing medical images. Multidisciplinary researchers have been working diligently for decades to develop automated diagnosis systems, and to this day it is one of the most active research areas \cite{schindelin2015imagej}. The task of a computer aided medical image analysis tool is twofold: segmentation and diagnosis. In the general Semantic Segmentation problem, the objective is partitioning an image into a set of non-overlapping regions, which allows the homogeneous pixels to be clustered together \cite{mcguinness2010comparative}. However, in the context of medical images the interest often lies in distinguishing only some interesting areas of the image, like the tumor regions \cite{codella2018skin}, organs \cite{yang2018autosegmentation} etc. This enables the doctors to analyze only the significant parts of the otherwise incomprehensible multimodal medical images \cite{naik2008automated}. Furthermore, often the segmented images are used to compute various features that may be leveraged in the diagnosis \cite{rouhi2015benign}. Therefore, image segmentation is of utmost importance and application in the domain of Biomedical Engineering.

Owing to the profound significance of medical image segmentation and the complexity associated with manual segmentation, a vast number of automated medical image segmentation methods have been developed, mostly focused on images of specific modalities. In the early days, simple rule-based approaches were followed; however, those methods failed to maintain robustness when tested on huge variety of data \cite{pham2000current}. Consequently, more adaptive algorithms were developed relying on geometric shape priors with tools of soft-computing \cite{mesejo2015biomedical} and fuzzy algorithms \cite{zheng2015image}. Nevertheless, these methods suffer from human biases and can not deal with the amount of variance in real world data.

Recent advancements in deep learning \cite{lecun2015deep} have shwon a lot of promises towards solving such issues. In this regard, Convolutional Neural Networks (CNN) \cite{lecun1998gradient} have been the most ground-breaking addition, which is dominating the field of Computer Vision. CNNs have been responsible for the phenomenal advancements in tasks like object classification \cite{krizhevsky2012imagenet}, object localization \cite{sermanet2013overfeat} etc., and the continuous improvements to CNN architectures are bringing further radical progresses \cite{simonyan2014very, szegedy2015going, he2016deep, szegedy2017inception}. Semantic Segmentation tasks have also been revolutionalized by Convolutional Networks. Since CNNs are more intuitive in performing object classification, Ciresan et al. \cite{ciresan2012deep} presented a sliding window based pipeline to perform semantic segmentation using CNN. Long et al. \cite{long2015fully} proposed a fully convolutional network (FCN) to perform end-to-end image segmentation, which surpassed the existing approaches. Badrinarayanan et al. \cite{badrinarayanan2015segnet} improved upon FCN, by developing a novel architecture namely, SegNet. SegNet consists of a 13 layer deep encoder network that extracts spatial features from the image, and a corresponding 13 layer deep decoder network that upsamples the feature maps to predict the segmentation masks. Chen et al. \cite{chen2018deeplab} presented DeepLap and performed semantic segmentation using atrous convolutions.

In spite of initiating a breakthrough in computer vision tasks, a major drawback of the CNN architectures is that they require massive volumes of training data. Unfortunately, in the context of medical images, not only the acquisition of images is expensive and complicated,   accurate annotation thereof adds even more to the complexity \cite{litjens2017survey}. Nevertheless, CNNs have shown great promise in medical image segmentation in recent years \cite{litjens2017survey,anwar2018medical}, and most of the credit go to U-Net \cite{ronneberger2015u}. The structure of U-Net is quite similar to SegNet, comprising an encoder and a decoder network. Furthermore, the corresponding layers of the encoder and decoder network are connected by skip connections, prior to a pooling and subsequent to a deconvolution operation respectively. U-Net has been showing impressive potential in segmenting medical images, even with a scarce amount of labeled training data, to an extent that it has become the de-facto standard in medical image segmentation \cite{litjens2017survey}. U-Net and U-Net like models have been successfully used in segmenting biomedical images of neuronal structures \cite{ronneberger2015u}, liver \cite{christ2016automatic}, skin lesion \cite{lin2017skin}, colon histology \cite{sirinukunwattana2017gland}, kidney \cite{cciccek20163d}, vascular boundary \cite{merkow2016dense}, lung nodule \cite{setio2017validation}, prostate \cite{yu2017volumetric}, etc. and the list goes on.

In this paper, in parallel to appreciating the capabilities of U-Net, the most popular and successful deep learning model for biomedical image segmentation, we diligently scrutinize the network architecture to discover some potential scopes of improvement. We argue and hypothesize that the U-Net architecture may be lacking in certain criteria and based on contemporary advancements in deep computer vision we propose some alterations to it. In the sequel, we develop a novel model called MultiResUNet, an enhanced version of U-Net, that we believe will significantly advance the state of the art in the domain of general multimodal biomedical image segmentation. We put our model to test using a variety of medical images originating from different modalities, and even with 3D medical images. From extensive experimentation with this diverse set of medical images, it was found that MultiResUNet overshadows the classical U-Net model in all the cases even with slightly less number of parameters. 

The contributions of this paper can be summarized as follows:

\begin{itemize}
    \item We analyze the U-Net model architecture in depth, and conjecture some potential opportunities for further enhancements
    \item Based on the probable scopes for improvement, we propose MultiResUNet, which is an enhanced version of the standard U-Net architecture.
    \item We experiment with different public medical image datasets of different modalities, and MultiResUNet shows superior accuracy.
    \item We also experiment with a 3D version of MultiResUNet, and it outperforms the standard 3D U-Net as well.
    \item Particularly, we examine some very challenging images and observe a significant improvement in using MultiResUNet over U-Net.
\end{itemize}

\section{Overview of the UNet Architecture}

Similar to FCN \cite{long2015fully} and SegNet \cite{badrinarayanan2015segnet}, U-Net \cite{ronneberger2015u} uses a network entirely of convolutional layers to perform the task of semantic segmentation. The network architecture is symmetric, having an \textit{Encoder} that extracts spatial features from the image, and a \textit{Decoder} that constructs the segmentation map from the encoded features. The \textit{Encoder} follows the typical formation of a convolutional network. It involves a sequence of two $3 \times 3$ convolution operations, which is followed by a max pooling operation with a pooling size of $2 \times 2$ and stride of 2. This sequence is repeated four times, and after each downsampling the number of filters in the convolutional layers are doubled. Finally, a progression of two $3 \times 3$ convolution operations connects the \textit{Encoder} to the \textit{Decoder}.

On the contrary, the \textit{Decoder} first up-samples the feature map using a $2 \times 2$ transposed convolution operation \cite{zeiler2010deconvolutional}, reducing the feature channels by half. Then again a sequence of two $3 \times 3$ convolution operations is performed. Similar to the \textit{Encoder}, this succession of up-sampling and two convolution operations is repeated four times, halving the number of filters in each stage. Finally, a $1 \times 1$ convolution operation is performed to generate the final segmentation map. All convolutional layers in this architecture except for the final one use the \textit{ReLU} (Rectified Linear Unit) activation function  \cite{lecun2015deep}; the final convolutional layer uses a \textit{Sigmoid} activation function.

\begin{figure}[h]
    \centering
    \includegraphics[width=\textwidth]{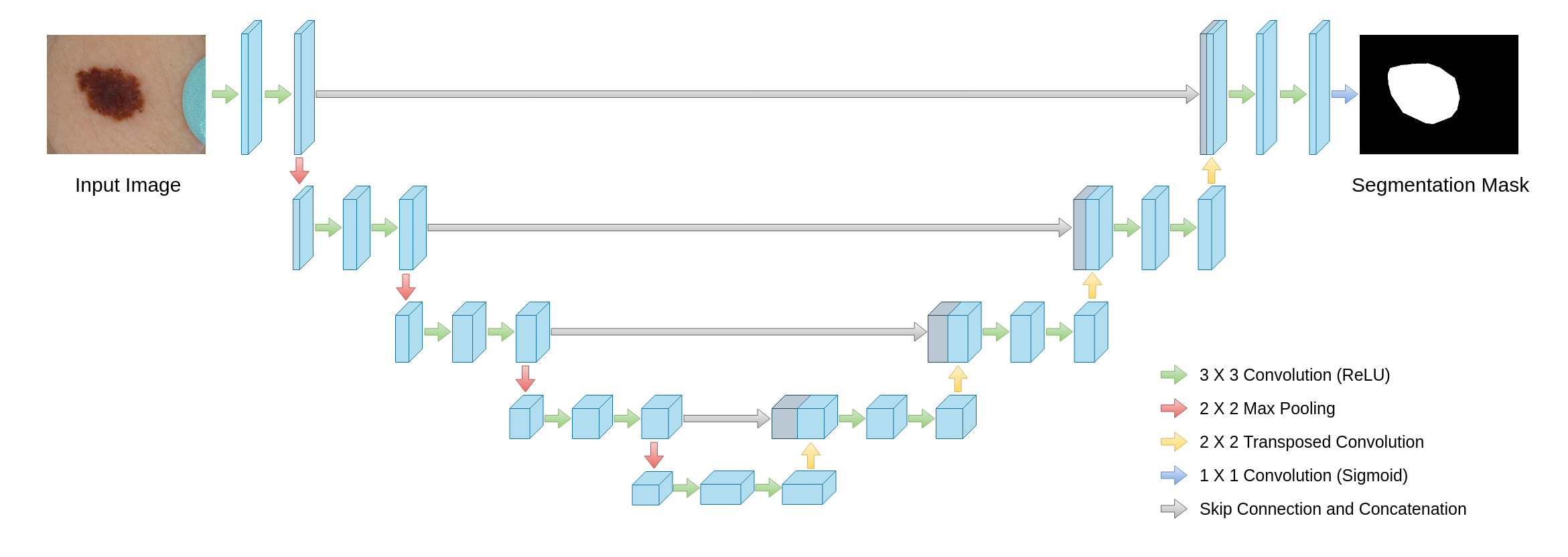}
    \caption{The U-Net Architecture. The model comprises an encoder and a decoder pathway, with skip connections between the corresponding layers.}
    \label{fig:unet}
\end{figure}

Perhaps, the most ingenious aspect of the U-Net architecture is the introduction of skip connections. In all the four levels, the output of the convolutional layer, prior to the pooling operation of the \textit{Encoder} is transfered to the \textit{Decoder}. These feature maps are then concatenated with the output of the upsampling operation, and the concatenated feature map is propagated to the successive layers. These skip connections allow the network to retrieve the spatial information lost by pooling operations \cite{drozdzal2016importance}. The network architecture is illustrated in Figure \ref{fig:unet}.

Subsequently, the U-Net architecture was extended through a few modifications to 3D U-Net for volumetric segmentation \cite{cciccek20163d}. In particular, the two dimensional convolution, max pooling, transposed convolution operations were replaced by their three dimensional counterparts. However, in order to limit the number of parameters, the depth of the network was reduced by one. Moreover, the number of filters were doubled before the pooling layers to avoid bottlenecks \cite{szegedy2016rethinking}. The original U-Net \cite{ronneberger2015u} did not use batch normalization \cite{ioffe2015batch}, however, they were experimented with in the 3D U-Net and astonishingly the results revealed that batch normalization may sometime even hurt the performance \cite{cciccek20163d}.

\section{Motivations and High Level Considerations}

U-Net has been a remarkable and the most popular Deep Network Architecture in medical imaging community, defining the state of the art in medical image segmentation \cite{drozdzal2016importance}. However, thorough contemplation of the U-Net architecture and drawing some parallels to the recent advancement of deep computer vision leads to some useful observations as described in the following subsections.

\subsection{Variation of Scale in Medical Images}
\label{multiresblock}
\begin{figure}[h]
    \centering
    \begin{subfigure}[b]{0.325\textwidth}
        \includegraphics[width=\textwidth]{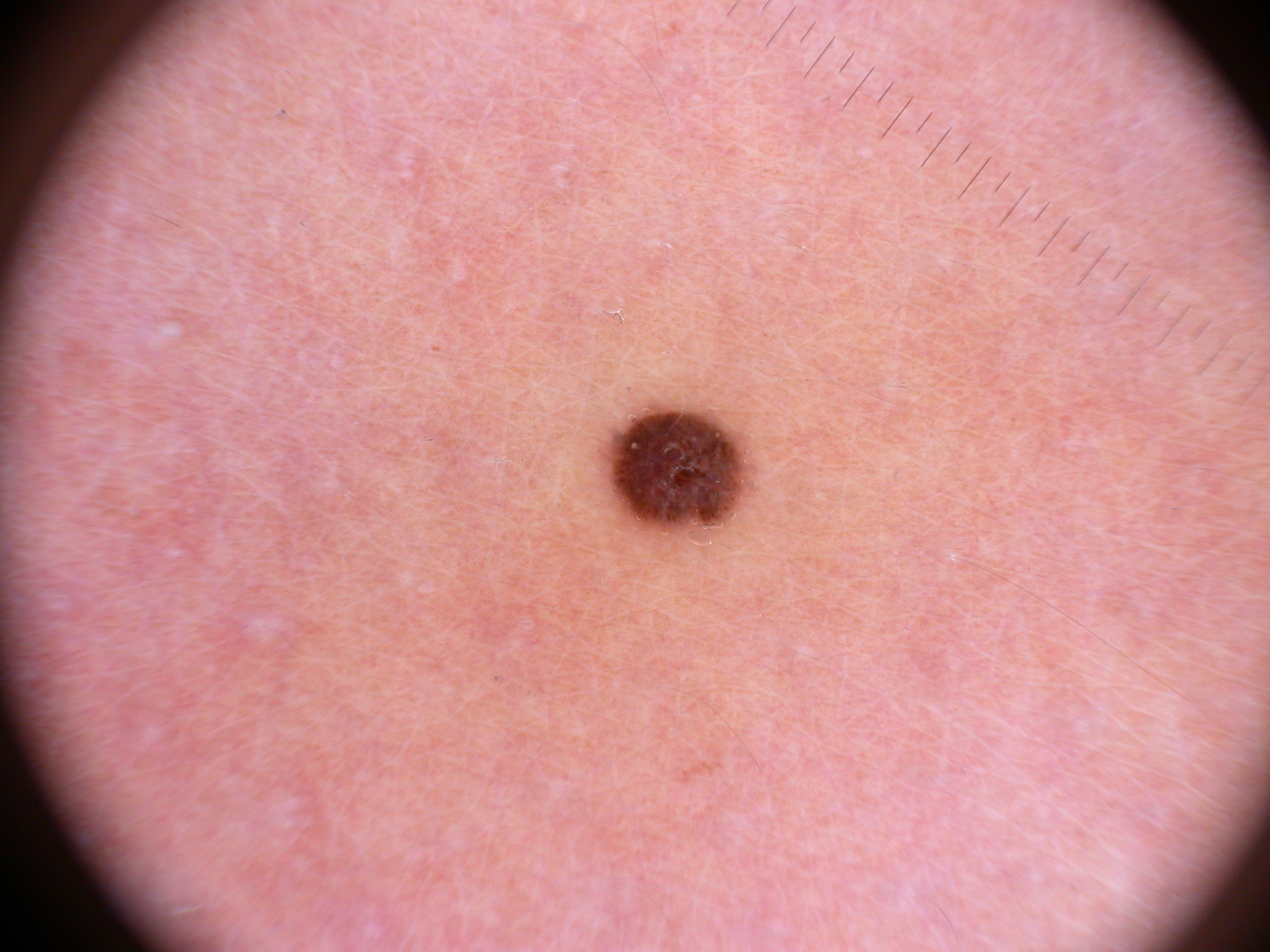}
        \caption{}
        \label{fig:skin_small}
    \end{subfigure}    
    \hfill
    \begin{subfigure}[b]{0.325\textwidth}
        \includegraphics[width=\textwidth]{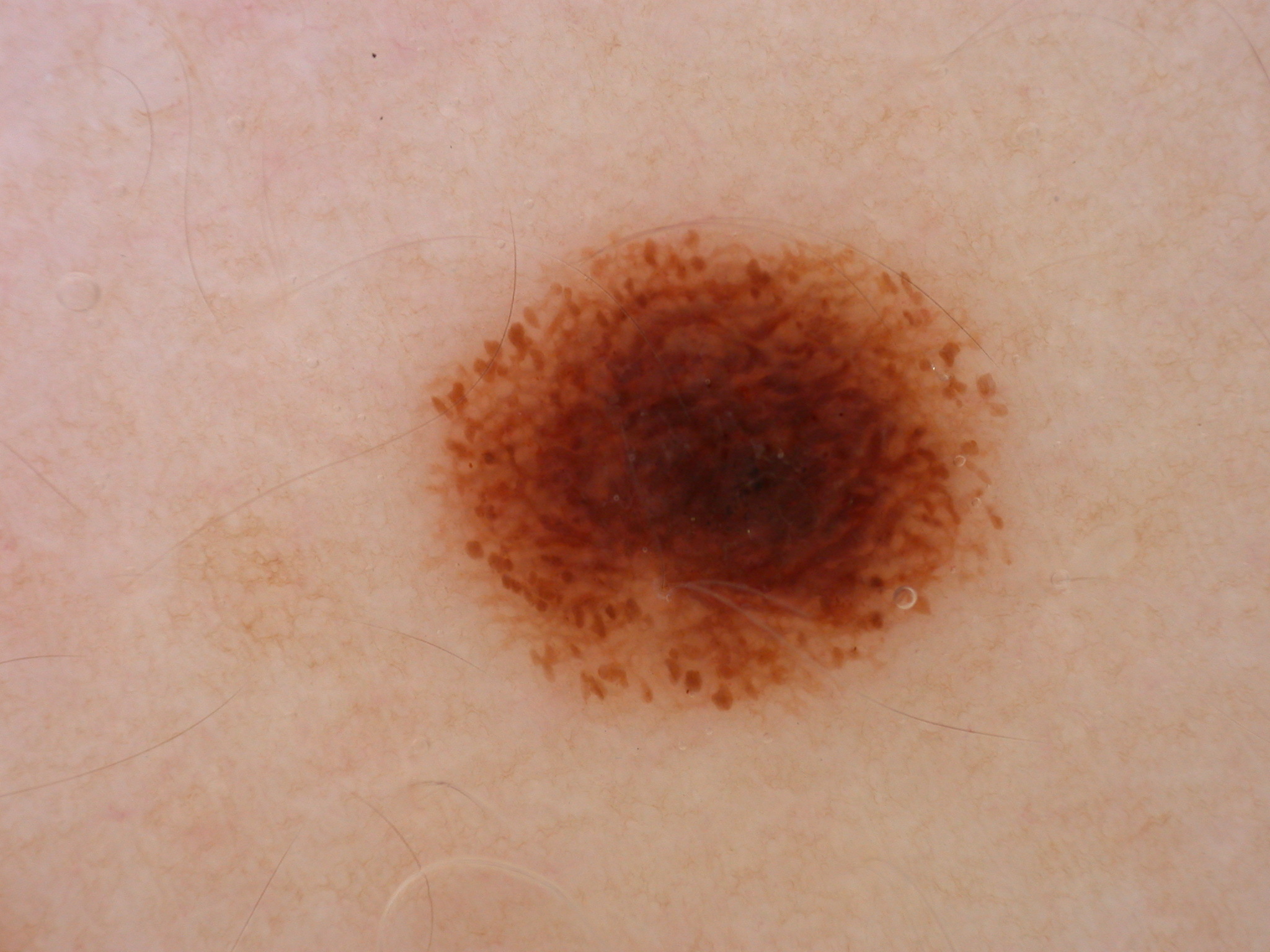}
        \caption{}
        \label{fig:skin_medium}
    \end{subfigure}
    \hfill
    \begin{subfigure}[b]{0.325\textwidth}
        \includegraphics[width=\textwidth]{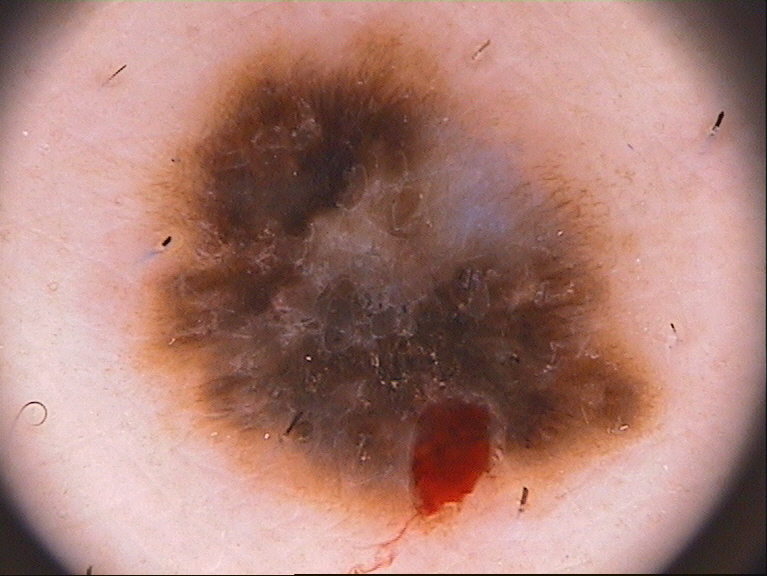}
        \caption{}
        \label{fig:skin_big}
    \end{subfigure}
    \caption{Variation of Scale in medical images. Fig. \ref{fig:skin_small}, \ref{fig:skin_medium}, \ref{fig:skin_big} are examples of dermoscopy images with small, medium and large size of lesions respectively. The images have been taken from the ISIC-2018 dataset.}
    \label{fig:scale_vary}
\end{figure}

In medical image segmentation, we are interested in segmenting cell necluei \cite{coelho2009nuclear}, organs \cite{yang2018autosegmentation}, tumors \cite{codella2018skin} etc. from images originating from various modalities. However, in most cases these objects of interest are of irregular and different scales. For example, in Figure \ref{fig:scale_vary} we have demonstrated that the scale of skin lesions can greatly vary in dermoscopy images. These situations frequently occur in different types medical image segmentation tasks.

%\begin{figure}[h]\centering\subfloat[]{\includegraphics[width=0.325\textwidth]{ISIC_0000125.jpg}\label{fig:skin_small}}\subfloat[]{\includegraphics[width=0.325\textwidth]{ISIC_0000108.jpg}\label{fig:skin_medium}}\subfloat[]{\includegraphics[width=0.325\textwidth]{ISIC_0000337.jpg}\label{fig:skin_big}}     \caption{Variation of Scale in medical images. Fig. \ref{fig:skin_small}, \ref{fig:skin_medium}, \ref{fig:skin_big} are examples of dermoscopy images with small, medium and large size of lesions respectively. The images have been taken from the ISIC-2018 dataset.}\label{fig:scale_vary}\end{figure}

Therefore, a network should be robust enough to analyze objects at different scales. Although this issue has been addressed in several deep computer vision works, to the best of our knowledge, this issue is still not addressed properly in the domain of medical image segmentation. Serre et al. \cite{serre2007robust} employed a sequence of fixed Gabor filters of varying scales to acknowledge the variation of scale in the image. Later on, the revolutionary Inception architecture \cite{szegedy2015going} introduced Inception blocks, that utilizes convolutional layers of varying kernel sizes in parallel to inspect the points of interest in images from different scales. These perceptions obtained at different scales are combined together and passed on deeper into the network.

\begin{figure}[h]
    \centering
    \begin{subfigure}[h]{0.23\textwidth}
        \includegraphics[width=\textwidth]{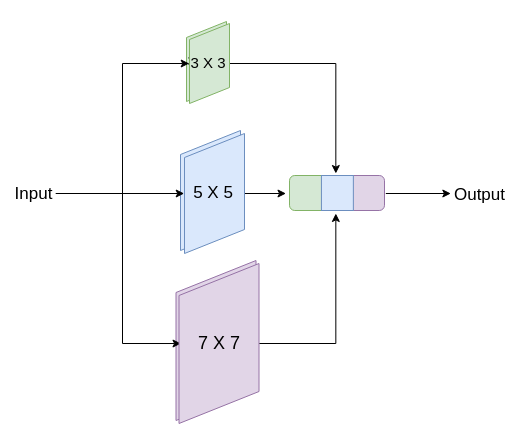}
        \caption{} 
        \label{fig:incpt1}
    \end{subfigure}
    \begin{subfigure}[h]{0.375\textwidth}
        \includegraphics[width=\textwidth]{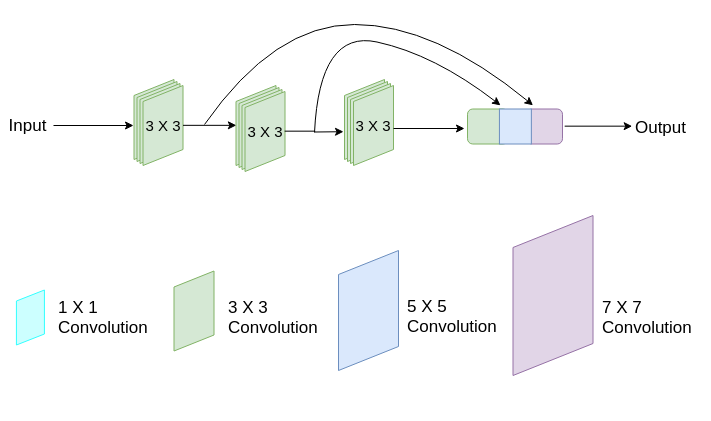}
        \caption{} 
        \label{fig:incpt2}
    \end{subfigure}
    \begin{subfigure}[h]{0.375\textwidth}
        \includegraphics[width=\textwidth]{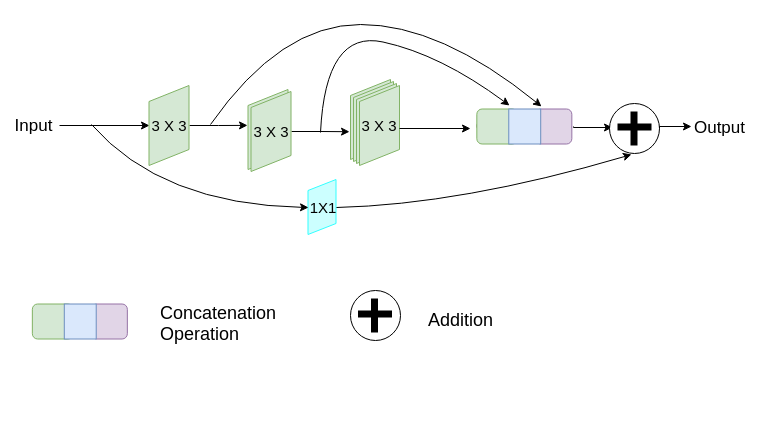} 
        \caption{} 
        \label{fig:incpt3}     
    \end{subfigure}
    \caption{Devolping the Proposed \textit{MultiRes} block. We start with a simple Inception like block by using $3\times3$, $5\times5$ and $7\times7$ convolutional filters in parallel and concatenating the generated feature maps (Fig. \ref{fig:incpt1}). This allows us to reconcile spatial features from different context size. Instead of using the $3\times3$, $5\times5$ and $7\times7$ filters in parallel, we factorize the bigger and more expensive $5\times5$ and $7\times7$ filters as a succession of $3\times3$ filters (Fig. \ref{fig:incpt2}) . Fig \ref{fig:incpt3} illustrates the \textit{MultiRes} block, where we have increased the number of filters in the successive three layers gradually and added a residual connection (along with $1\times1$ filter for conserving dimensions).}
    \label{fig:incpt23}
\end{figure}

In the U-Net architecture, after each pooling layer and transposed convolutional layer a sequence of two $3 \times 3$ convolutional layers are used. As explained in \cite{szegedy2016rethinking}, this series of two $3 \times 3$ convolutional operation actually resembles a $5 \times 5$ convolutional operation. Therefore, following the approach of Inception network, the simplest way to augment U-Net with multi-resolutional analysis is to incorporate $3 \times 3$, and $7 \times 7$ convolution operations in parallel to the $5 \times 5$ convolution operation, as shown in Figure \ref{fig:incpt1}.

%\begin{figure}[h]
%    \centering
%    \includegraphics[width=0.5\textwidth]{incpt1.png}
%    \caption{Simple Inception like block. Here we have used $3\times3$, $5\times5$ and $7\times7$ convolutional filters in parallel and concatenated the generated feature maps. This allows us to reconcile spatial features from different context size.}
%    \label{fig:incpt1}
%\end{figure}

Therefore, replacing the convolutional layers with Inception-like blocks should facilitate the U-Net architecture to reconcile the features learnt from the image at different scales. Another possible option is to use strided convolutions \cite{wang2018understanding}, but in our experiments it was overshadowed by U-Net using Inception-like blocks. Despite the gain in performance, the introduction of additional convolutional layers in parallel extravagantly increases the memory requirement. Therefore, we improvise with the following ideas borrowed from \cite{szegedy2016rethinking}. We factorize the bigger, more demanding $5 \times 5$ and $7 \times 7$ convolutional layers, using a sequence of smaller and lightweight $3 \times 3$ convolutional blocks, as shown in Figure \ref{fig:incpt2}. The outputs of the 2nd and 3rd $3 \times 3$ convolutional blocks effectively approximate the $5 \times 5$ and $7 \times 7$ convolution operations respectively. We hence take the outputs from the three convolutional blocks and concatenate them together to extract the spatial features from different scales. From our experiments, it was seen that the results of this compact block closely resemble that of the memory intensive Inception-like block decsribed earlier. This outcome is in line with the findings of \cite{szegedy2016rethinking}, as the adjacent layers of a vision network are expected to be correlated. 

Despite that this modification greatly reduces the memory requirement, it is still quite demanding. This is mostly due to the fact that in a deep network if two convolutional layers are present in a succession, then the number of filters in the first one has a quadratic effect over the memory \cite{szegedy2015going}. Therefore, instead of keeping all the three consecutive convolutional layers of equal number of filters, we gradually increase the filters in those (from 1 to 3), to prevent the memory requirement of the earlier layers from exceedingly propagating to the deeper part of the network. We also add a residual connection because of their efficacy in biomedical image segmentaion \cite{drozdzal2016importance}, and for the introduction of the $1 \times 1$ convolutional layers, which may allow us to comprehend some additional spatial information. We call this arrangement a `\textit{MultiRes block}', as shown in Figure \ref{fig:incpt3}.

\subsection{Probable Semantic Gap between the Corresponding Levels of Encoder-Decoder}
\label{sec:respath}
An ingenious contribution of the U-Net architecture was the introduction of shortcut connections between the corresponding layers before and after the max-pooling and the deconvolution layers respectively. This enables the network to propagate from encoder to decoder, the spatial information that gets lost during the pooling operation.

Despite preserving the dissipated spatial features, a flaw of the skip connections may be speculated as follows. For instance, the first shortcut connection bridges the encoder before the first pooling with the decoder after the last deconvolution operation. Here, the features coming from the encoder are supposed to be lower level features as they are computed in the earlier layers of the network. On the contrary, the decoder features are supposed to be of much more higher level, since they are computed at the very deep layers of the network. Therefore, they go through more processing. Hence, we observe a possible semantic gap between the two sets of features being merged. We conjecture that the fusion of these two arguably incompatible sets of features could cause some discrepancy throughout the learning and thus adversely affect the prediction procedure. It may be noted that, the amount of discrepancy is likely to decrease gradually as we move towards the succeeding shortcut connections. This can be attributed to the fact that, not only the features from the encoder are going through more processing, but also we are fusing them with decoder features of much juvenile layers.

\begin{figure}[h]
    \centering
    \includegraphics[width=\textwidth]{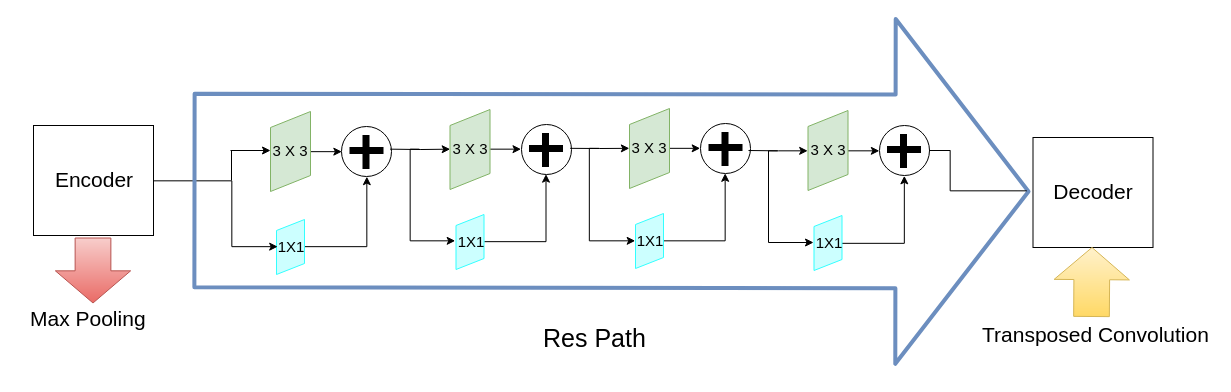}
    \caption{Proposed \textit{Res} path. Instead of combining the the encoder feature maps with the decoder feature in a straight-forward manner, we pass the encoder features through a sequence of convolutional layers. These additional non-linear operations are expected to reduce the semantic gap between encoder and decoder features. Furthermore, residual connections are also introduced as they make the learning easier and are very useful in deep convolutional networks.}
    \label{fig:respath}
\end{figure}

Therefore, to alleviate the disparity between the encoder-decoder features, we propose to incorporate some convolutional layers along the shortcut connections. Our hypothesis is that these additional non-linear transformations on the features propagating from the encoder stage should account for the further processing done during the by decoder stage therein. Furthermore, instead of using the usual convolutional layers we introduce residual connections to them as they make the learning easier \cite{szegedy2017inception} and is proven to be having great potential in medical image analysis \cite{drozdzal2016importance}. This idea is inspired from the image to image conversion using convolutional neural networks \cite{mao2016image}, where pooling layers are not favorable for the loss of information. Thus, instead of simply concatenating the feature maps from the encoder stage to the decoder stage, we first pass them through a chain of convolutional layers with residual connections, and then concatenate with the decoder features. We call this proposed shortcut path `\textit{Res path}', illustrated in Fig. \ref{fig:respath}. Specifically, $3\times3$ filters are used in the convolutional layers and $1\times1$ filters accompany the residual connections.

\section{Proposed Architecture}

%Considering the probable shortcomings of the U-Net architecture, we propose a novel architecture MultiResUNet, augmented with \textit{MultiRes blocks} and \textit{Res paths}.

In the MultiResUNet model, we replace the sequence of two convolutional layers with the proposed \textit{MultiRes} block as introduced in Section \ref{sec:respath}. For each of the \textit{MultiRes} blocks, we assign a parameter $W$, which controls the number of filters of the convolutional layers inside that block. To maintain a comparable relation between the numbers of parameters in original U-Net and the proposed model, we compute the value of $W$ as follows:

\begin{equation}
    W = \alpha \times U
\end{equation}

Here, $U$ is the number of filters in the corresponding layer of the U-Net and $\alpha$ is a scaler coefficient. Decomposing $W$ to $U$ and $\alpha$ provides a convenient way to both controlling the number of parameters and keeping them comparable to U-Net. We compare our proposed model with an U-Net, having $\#filters = [32,64,128,256,512]$ along the levels, which are also the values of $U$ in our model. We selected $\alpha=1.67$ as it keeps the number of parameters in our model slightly below that of the U-Net.

In Section \ref{sec:respath}, we pointed out that it is beneficial to gradually increase the number of filters in the successive convolutional layers inside a \textit{MultiRes} block, instead of keeping them the same. Hence, we assign $\floor*{\frac{W}{6}}$, $\floor*{\frac{W}{3}}$ and $\floor*{\frac{W}{2}}$ filters to the three successive convolutional layers respectively, as this combination achieved the best results in our experiments. Also it can be noted that similar to the U-Net architecture, after each pooling or deconvolution operation the value of $W$ gets doubled.

In addition to introducing the \textit{MultiRes} blocks, we also replace the ordinary shortcut connections with the proposed $Res$ paths. Therefore, we apply some convolution operations on the feature maps propagating from the encoder stage to the decoder stage. In Section \ref{multiresblock}, we hypothesized that the intensity of the semantic gap between the encoder and decoder feature maps are likely to decrease as we move towards the inner shortcut paths. Therefore, we also gradually reduce the number of convolutional blocks used along the $Res$ paths. In particular, we use $4,3,2,1$ convolutional blocks respectively along the four $Res$ paths. Also, in order to account for the number of feature maps in encoder-decoder, we use $32,64,128,256$ filters in the blocks of the four $Res$ paths respectively.

\begin{figure}[h]
    \centering
    \includegraphics[width=\textwidth]{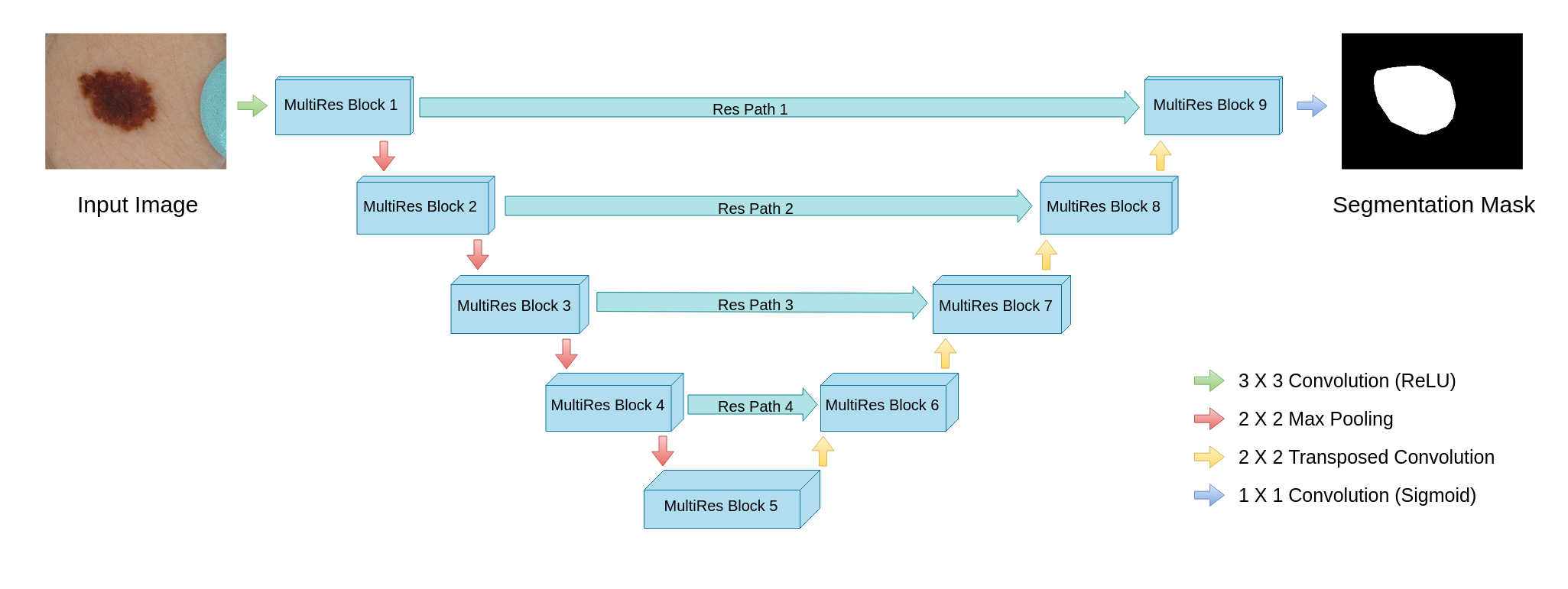}
    \caption{Proposed MultiResUNet architecture. We replace the sequences of two convolutional layers in the U-Net architectures with the proposed \textit{MultiRes} block. Furthermore, instead of using plain shortcut connections, we use the proposed \textit{Res} paths.}
    \label{fig:mresunet}
\end{figure}

All the convolutional layers except for the output layer, used in this network is activated by the \textit{ReLU} (Rectified Linear Unit) activation function \cite{lecun2015deep}, and are batch-normalized \cite{ioffe2015batch}. Similar to the U-Net model, the output layer is activated by a \textit{Sigmoid} activation function. We present a diagram of the proposed MultiResUNet model in Fig. \ref{fig:mresunet}. The architectural details are described in Table \ref{table:mresunet_archi}. 

\begin{table}[h]\caption{MultiResUNet Architecture Details}\label{table:mresunet_archi}
\begin{tabular}{|c|c|c|c|c|c|}\hline\multicolumn{6}{|c|}{MultiResUNet}\\\hline Block&Layer(filter size)&\#filters&Path&Layer(filter size)&\#filters\\\hline\multirow{4}{*}{\begin{tabular}[c]{@{}c@{}}MultiRes Block 1\\\\MultiRes Block 9\end{tabular}}&Conv2D(3,3)&8&\multirow{8}{*}{Res Path 1}&Conv2D(3,3)&32\\\cline{2-3}\cline{5-6}&Conv2D(3,3)&17&&Conv2D(1,1)&32\\\cline{2-3}\cline{5-6}&Conv2D(3,3)&26&&Conv2D(3,3)&32\\\cline{2-3}\cline{5-6}&Conv2D(1,1)&51&&Conv2D(1,1)&32\\\cline{1-3}\cline{5-6}\multirow{4}{*}{\begin{tabular}[c]{@{}c@{}}MultiRes Block 2\\\\MultiRes Block 8\end{tabular}}&Conv2D(3,3)&17&&Conv2D(3,3)&32\\\cline{2-3}\cline{5-6}&Conv2D(3,3)&35&&Conv2D(1,1)&32\\\cline{2-3}\cline{5-6}&Conv2D(3,3)&53&&Conv2D(3,3)&32\\\cline{2-3}\cline{5-6}&Conv2D(1,1)&105&&Conv2D(1,1)&32\\\hline\multirow{4}{*}{\begin{tabular}[c]{@{}c@{}}MultiRes Block 3\\\\MultiRes Block 7\end{tabular}}&Conv2D(3,3)&35&\multirow{6}{*}{Res Path 2}&Conv2D(3,3)&64\\\cline{2-3}\cline{5-6}&Conv2D(3,3)&71&&Conv2D(1,1)&64\\\cline{2-3}\cline{5-6}&Conv2D(3,3)&106&&Conv2D(3,3)&64\\\cline{2-3}\cline{5-6}&Conv2D(1,1)&212&&Conv2D(1,1)&64\\\cline{1-3}\cline{5-6}\multirow{4}{*}{\begin{tabular}[c]{@{}c@{}}MultiRes Block 4\\\\MultiRes Block 6\end{tabular}}&Conv2D(3,3)&71&&Conv2D(3,3)&64\\\cline{2-3}\cline{5-6}&Conv2D(3,3)&142&&Conv2D(1,1)&64\\\cline{2-6}&Conv2D(3,3)&213&\multirow{4}{*}{ResPath 3}&Conv2D(3,3)&128\\\cline{2-3}\cline{5-6}&Conv2D(1,1)&426&&Conv2D(1,1)&128\\\cline{1-3}\cline{5-6}\multirow{4}{*}{MultiRes Block 5}&Conv2D(3,3)&142&&Conv2D(3,3)&128\\\cline{2-3}\cline{5-6}&Conv2D(3,3)&284&&Conv2D(1,1)&128\\\cline{2-6}&Conv2D(3,3)&427&\multirow{2}{*}{Res Path 4}&Conv2D(3,3)&256\\\cline{2-3}\cline{5-6}&Conv2D(1,1)&853&&Conv2D(1,1)&256\\\hline\end{tabular}\end{table}

\section{Datasets}
\label{sec:dataset}
Curation of medical imaging datasets are challenging compared to the traditional computer vision datasets. Expensive imaging equipments, sophisticated image acquisition pipelines, necessity of expert annotation, issues of privacy-all adds to the complexity of developing medical imaging datasets \cite{litjens2017survey}. As a result, only a few public medical imaging benchmark datasets exist, and they only contain a handful of images each. In order to assess the efficacy of the proposed architecture, we tried to evaluate it on a variety of image modalities. More specifically, we selected datasets that are as heterogeneous as possible from each other. Also, each of these datasets poses a unique challenge of their own (more details are given in Section \ref{sec:results} and Section \ref{sec:discussions}). The datasets used in the experiments are briefly described below (also see Table \ref{tab:dataset} for an overview).

\begin{table}[h]
\caption{Overview of the Datasets.}
\label{tab:dataset}
\centering
\small
\begin{tabular}{|c|c|c|c|c|}
\hline
Modality & Dataset & No. of images & Original Resolution & Input Resolution \\ \hline
Fluorescence Microscopy & Murphy Lab & 97 & Variable & $256 \times 256$ \\ \hline
Electron Microscopy & ISBI-2012 & 30 & $512 \times 512$ & $256 \times 256$ \\ \hline
Dermoscopy & ISIC-2018 & 2594 & Variable & $256 \times 192$ \\ \hline
Endoscopy & CVC-ClinicDB & 612 & $384 \times 288$  & $256 \times 192$ \\ \hline
MRI & BraTS17 & \begin{tabular}[c]{@{}c@{}}210 HGG + \\ 75 LGG\end{tabular} & $240 \times 240 \times 155$ & $80 \times 80 \times 48$ \\ \hline
\end{tabular}
\end{table}

%In contrast to the traditional computer vision problems, medical image analysis presents some unique challenges. For instance, to curate a dataset of images for object classification or localization, ordinary cameras are sufficient. But in order to develop a database of medical images expensive imaging devices are required and a sophisticated pipeline needs to be followed \cite{hendee2003medical}. Even after such a complicated image acquisition procedure, the complexity still persists in labelling the data, as domain experts are essential to label the data properly. Moreover, labelling the data is also to an extent subjective among the medical experts, which results in a lot of noise in the data. To 

\subsection{Fluorescence Microscopy Image}
We used the fluorescence microscopy image dataset developed by Murphy Lab \cite{coelho2009nuclear}. This dataset contains 97 fluorescence microscopy images and a total of 4009 cells are contained in these images. Half of the cells are U2OS cells and the other half comprises NIH3T3 cells. The nuclei were segmented manually by experts. The nuclei are irregular in terms of brigtness and the imags often contain noticeable debris, making this a challenging dataset of bright-field microscopy images. The original resolution of the images range from $1349 \times 1030$ to $1344 \times 1024$; they have been resized to $256 \times 256$ for computational constraints.

\subsection{Electron Microscopy Image}

To observe the effectiveness of the architecture with electron microscopy images, we used the dataset of the ISBI-2012 : 2D EM segmentation challenge \cite{arganda2015crowdsourcing,cardona2010integrated}. This dataset contains only 30 images from a serial section Transmission Electron Microscopy (ssTEM) of the Drosophila first instar larva ventral nerve cord \cite{cardona2010integrated}. The images face slight alignment errors, and are corrupted with noises. The resolution of the images is $512 \times 512$, but they have been resized to $256 \times 256$ due to computational limitations.

\subsection{Dermoscopy Image}

We acquired the dermoscopy images from the ISIC-2018 : Lesion Boundary Segmentation challenge dataset. The data for this challenge were extracted from the ISIC-2017 dataset \cite{codella2018skin} and the HAM10000 dataset  \cite{tschandl2018ham10000}. The compiled dataset contains a total of 2594 images of different types of skin lesions with expert annotation. The images are of various resolutions, but they have all been resized to $256 \times 192$, maintaining the average aspect ratio, for computational purposes.

\subsection{Endoscopy Image}

We used the CVC-ClinicDB \cite{bernal2015wm}, a colonoscopy image database for our experiments with endoscopy images. The images of this dataset were extracted from frames of 29 colonoscopy video sequences. Only the images with polyps were considered, resulting in a total of 612 images. The images are originally of resolution $384 \times 288$, but have been resized to $256 \times 192$, maintaining the aspect ratio.

\subsection{Magnetic Resonance Image}
All the datasets described previously contains 2D medical images. In order to evaluate our proposed architecture with 3D medical images, we used the magnetic resonance images (MRI) from the BraTS17 competition database \cite{menze2015multimodal,bakas2017advancing}. This dataset contains 210 glioblastoma (HGG) and 75 lower grade glioma (LGG) multimodal MRI scans. These multimodal scans include native (T1), post-contrast T1-weighted (T1Gd), T2-weighted (T2) and T2 Fluid Attenuated Inversion Recovery (FLAIR) volumes, which were acquired following different clinical protocols and various scanners from 19 institutions. The images are of dimensions $240 \times 240 \times 155$ but have been resized to $80 \times 80 \times 48$ for computational ease. All the four modalities, namely, T1, T1Gd, T2 and FLAIR are used as four different channels in evaluating the 3D variant of our model.

\section{Experiments}
\label{sec:experiments}
We used Python, more specifically Python3 programming language to conduct the experiments \cite{van2007python}. The network models were implemented using Keras \cite{chollet2015keras} with Tensorflow backend \cite{abadi2016tensorflow}. Our model implementation is available in the following github repository:

\centerline{\url{https://github.com/nibtehaz/MultiResUNet}}

The experimets were conducted in a desktop computer with intel core i7-7700 processor (3.6 GHz, 8 MB cache) CPU, 16 GB RAM, and NVIDIA TITAN Xp (12 GB, 1582 MHz) GPU.

\subsection{Baseline Model}

Since the proposed architecture, MultiResUNet, is targeted towards improving the state of the art U-Net architecture for medical image segmentation, we compared its performance with the U-Net architecture as the baseline. %It is worth to note that U-Net has outperformed FCN \cite{long2015fully}, SegNet\cite{badrinarayanan2015segnet} architectures for medical image segmentation.%
To keep the number of parameters comparable to our proposed MultiResUNet, we implemented the original U-Net \cite{badrinarayanan2015segnet} having five layer deep encoder and decoder, with filter numbers of 32, 64, 128, 256, 512. %The number of parameters of this network is comparable to our proposed , as shown in Table \ref{tab:param}.

Also, as the baseline for 3D image segmentation we used the 3D counterpart of the U-Net as described in the original paper \cite{cciccek20163d}. The 3D version of the MultiResUNet is constructed simply by substituting the 2D convolutional layers, pooling layers and transposed convolution layers, with their 3D variants respectively, without any further alterations. 

The number of parameters of the models are presented in Table \ref{tab:param}. In both the cases the proposed networks require slightly lesser number of parameters.

\begin{table}[h]
\caption{Models used in our experiments}
\label{tab:param}
\begin{tabular}{|c|c|c|c|}
\hline
\multicolumn{2}{|c|}{2D}                            & \multicolumn{2}{c|}{3D}                               \\ \hline
Model                                  & Parameters & Model                                    & Parameters \\ \hline
\multicolumn{1}{|l|}{U-Net (baseline)} & 7,759,521  & \multicolumn{1}{l|}{3D U-Net (baseline)} & 19,078,593 \\ \hline
MultiResUNet (proposed)                & 7,262,750  & MultiResUNet 3D (proposed)               & 18,657,689 \\ \hline
\end{tabular}
\end{table}

\subsection{Pre-processing / Post-processing}

The objective of the experiments is to investigate the superiority of the proposed MultiResUNet architecture over the original U-Net as a general model. Therefore, no domain specific pre-processing was performed. The only pre-processing the input images went through was that they were resized to fit into the GPU memory and the pixel values were divided by 255 to bring them to the $[0 \dots 1]$ range. Similarly, no application specific post-processing was performed. Since, the final layer is activated by a Sigmoid function, it produces outputs in the range $[0 \dots 1]$. Therefore, we applied a threshold of 0.5 to obtain the segmentation map of the input images.

\subsection{Training Methodology}

The task of semantic segmentation is to predict the individual pixels whether they represent a point of interest, or are merely a part of the background. Therefore, this problem ultimately reduces to a pixel-wise binary classification problem. Hence, as the loss function of the network we simply took the binary cross-entropy function and minimized it.

Let, for an image $X$, the ground truth segmentation mask is $Y$, and the segmentation mask predicted by the model is $\hat{Y}$. For a pixel $px$, the network predicts $\hat{y}_{px}$, whereas, the ground truth value is $y_{px}$. The binay cross-entropy loss for that image is defined as:

\begin{equation}
    \textit{Cross Entropy}(X,Y,\hat{Y}) = \sum_{px \in X}-(y_{px} \log(\hat{y}_{px}) + (1-y_{px}) \log(1-\hat{y}_{px}))
\end{equation}

For a batch containing $n$ images the loss function $J$ becomes,

\begin{equation}
    J = \frac{1}{n} \sum_{i=1}^{n}\textit{Cross Entropy}(X_i,Y_i,\hat{Y}_i)
\end{equation}

We minimized the binary cross-entropy loss, hence trained the model using the Adam optimizer \cite{kingma2014adam}. Adam adaptively computes different learning rates for different parameters from estimates of first and second moments of the gradients. This idea, in fact combines the advantages of both AdaGrad \cite{duchi2011adaptive} and RMSProp \cite{tieleman2012lecture}; therefore Adam has been often used in benchmarking deep learning models as the default choice \cite{ruder2016overview}. Adam has a number of parameters including $\beta_1$ and $\beta_2$ which control the decay of first and second moment respectively. However, in this work we used Adam with the parameters mentioned in the original paper. The models were trained for $150$ epochs using Adam optimizer. The reason of selecting $150$ as the number of epochs is due to the fact that after $150$ epochs neither of the models were showing any further improvements.

\subsection{Evaluation Metric}

In semantic segmentation, usually the points of interest comprise a small segment of the entire image. Therefore, metrics like precision, recall are inadequate and often lead to false sense of superiority, inflated by the perfection of detecting the background. Hence, Jaccard Index has been widely used to evaluate and benchmark image segmentation and object localization algorithms \cite{mcguinness2010comparative}. Jaccard Index for two sets $A$ and $B$ are defined as the ratio of the intersection and union of the two sets:

\begin{equation}
    \text{\textit{Jaccard Index}} = \frac{Intersection}{Union} = \frac{A \cap B}{A \cup B}
\end{equation}

In our case, the set $A$ represents the ground truth binary segmentation mask $Y$, and set $B$ corresponds to the predicted binary segmentation mask $\hat{Y}$. Therefore, by taking the Jaccard Index as the metric, we not only emphasize on precise segmentation, but also penalize under-segmentation and over-segmentation.

\subsection{$k$-Fold Cross Validation}

Cross-Validation tests estimate the general effectiveness of an algorithm on an independent dataset, ensuring a balance between bias and variance. In a $k$-Fold cross-validation test, the dataset $D$ is randomly split into $k$ mutually exclusive subsets $D_1,D_2, \cdots , D_k$ of equal or near equal size \cite{kohavi1995study}. The algorithm is run $k$ times subsequently, each time taking one of the $k$ splits as the validation set and the rest as the training set.
In order to evaluate the segmentation accuracy of both the baseline U-Net and proposed MultiResUNet architecture, we performed 5-Fold Cross Validation tests on each of the different datasets. 

Since, this is a deep learning pipeline, the best performing result on the validation set achieved through the total number of epochs (150 in our case) performed is recorded in each run. Finally, combining the results of all the $k$ runs gives an overall estimation of the performance of the algorithm.

\section{Results}
\label{sec:results}
\subsection{MultiResUNet Consistently Outperforms U-Net}
%\subsection{MultiResUNet Consistently Achieves Higher Scores for Multimodal Medical Images}

As described in Section \ref{sec:dataset} and Section \ref{sec:experiments} to evaluate the performance of the proposed architecture, we performed experiments with diversified classes of medical images, each with a unique challenge of its own. In particular, we have performed 5-fold cross validation and observed the performance of our proposed MultiResUNet and the baseline, U-Net. In each run the best results obtained on the validation set through the 150 epochs performed was noted and they were summarised from the 5 runs to obtain the final result.

The results of the 5-Fold Cross Validation for both the proposed MultiResUNet model and baseline U-Net model on the different datasets are presented in Table \ref{table:comparison}. It should be noted that for better readability the fractional Jaccard Index values have been converted to percentage ratios $(\%)$.

\begin{table}[h]
\caption{Results of 5-fold cross validation. Here, we present the best obtained results in the five folds, of both U-Net and MultiResUNet, for all the datasets used. We also mention the relative improvement of MultiResUNet over U-Net. It should be noted that, for better readability the fractional values of Jaccard Index have been converted to percentage ratios (\%).}
\label{table:comparison}
\centering
\begin{tabular}{|c|c|c|c|}
\hline
Modality                & MultiResUNet $(\%)$    & U-Net $(\%)$            & Relative Improvement $(\%)$ \\ \hline
Dermoscopy              & 80.2988 ± 0.3717 & 76.4277 ± 4.5183 & 5.065      \\ \hline
Endoscopy               & 82.0574 ± 1.5953 & 74.4984 ± 1.4704 & 10.1465       \\ \hline
Fluorescence Microscopy & 91.6537 ± 0.9563 & 89.3027 ± 2.1950 & 2.6326       \\ \hline
Electron Microscopy     & 87.9477 ± 0.7741 & 87.4092 ± 0.7071 & 0.6161      \\ \hline
MRI                     & 78.1936 ± 0.7868 & 77.1061 ± 0.7768 &      1.4104       \\ \hline
\end{tabular}
\end{table}

From the table, It can be observed that our proposed model outperforms the U-Net architecture in segmenting all different types of medical images. Most notably, remarkable improvements are observed for Dermoscopy and Endoscopy images. These images tend to be a bit less uniform and often they appear confusing even to a trained eye (more details are discussed in a later section). Therefore, this improvement is of great significance. For Fluorescence Microscopy images our model also achieve a $2.6326\%$ relative improvement over U-Net, and despite having a lesser number of parameters, still it achieves a relative improvement of $1.4104\%$ for MRI images. Only for Electron Microscopy images U-Net seems to be on par with our proposed model, yet in that case the latter obtains slightly better results (relative improvement of $0.6161\%$).

\subsection{MultiResUNet can Obtain Better Results in Less Number of Epochs}

\begin{figure}[h]
    \centering
    \begin{subfigure}[h]{0.325\textwidth}
        \includegraphics[width=\textwidth]{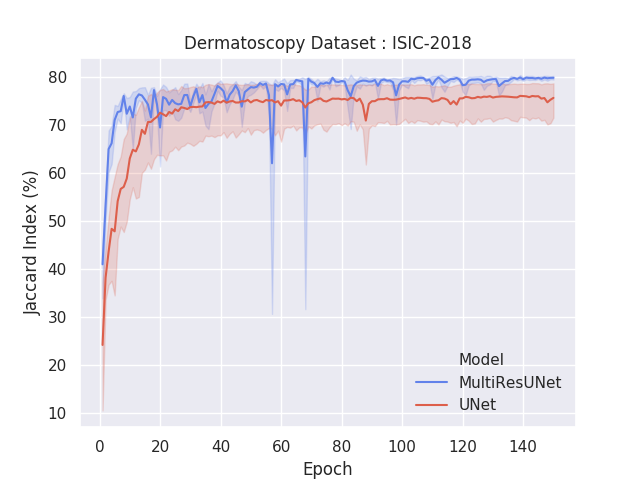}
        \caption{Dermatoscopy}
        \label{fig:epoch_derm}
    \end{subfigure}
    \begin{subfigure}[h]{0.325\textwidth}
        \includegraphics[width=\textwidth]{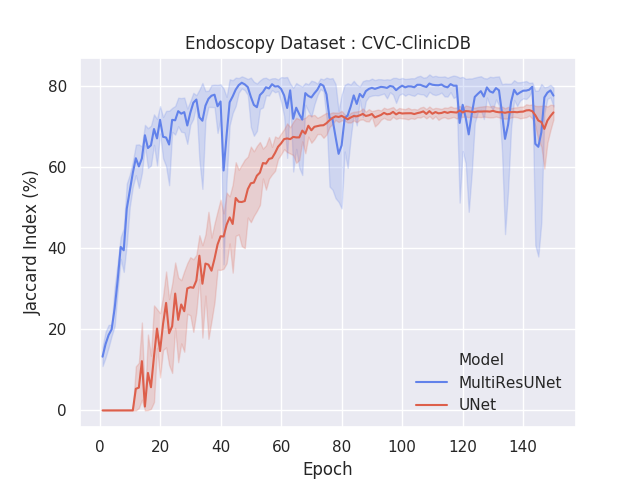}
        \caption{Endoscopy}
        \label{fig:epoch_endo}
    \end{subfigure}
    \begin{subfigure}[h]{0.325\textwidth}
        \includegraphics[width=\textwidth]{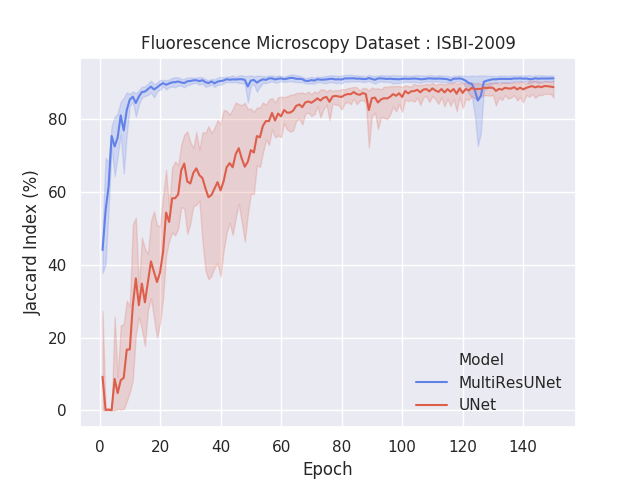}
        \caption{Fluorescence Microscopy}
        \label{fig:epoch_flu}
    \end{subfigure}
    \begin{subfigure}[h]{0.325\textwidth}
        \includegraphics[width=\textwidth]{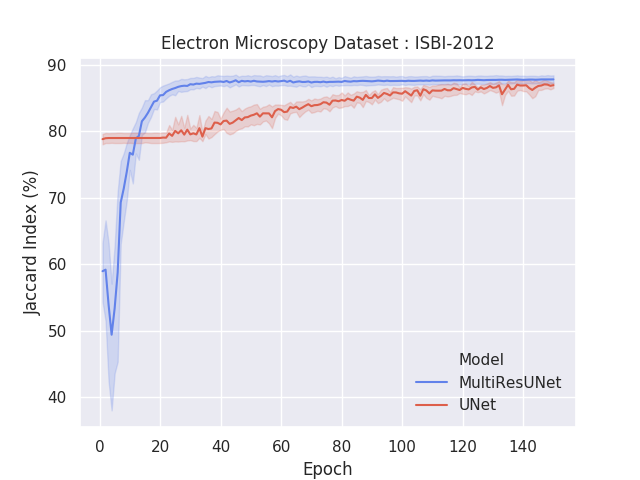}
        \caption{Electron Microscopy}
        \label{fig:epoch_elec}
    \end{subfigure}
    \begin{subfigure}[h]{0.325\textwidth}
        \includegraphics[width=\textwidth]{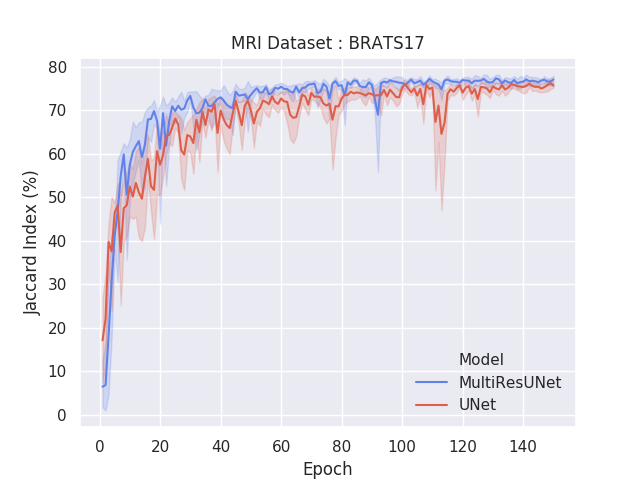}
        \caption{MRI}
        \label{fig:epoch_mri}
    \end{subfigure}
    \caption{Progress of the validation performance with the number of epochs. We record the value of Jacard Index on validation data after each epoch. It can be observed that not only MultiResUNet outperforms the U-Net model, but also the standard deviation of MultiResUNet is much smaller.}
    \label{fig:epochs}
\end{figure}

In addition to analyzing the best performing models from each run, we also monitored how the model performance progressed with epochs. In Figure \ref{fig:epochs}, the performance on the validation data on each epoch is shown, for all the datasets. We have presented the band of Jaccard Index values at a certain epoch in the 5-fold cross validation. It can be noted that for all the cases our proposed model attains convergance much faster. This can be attributed to the synergy between residual connections and batch normalization \cite{drozdzal2016importance}. Moreover, apart from Fig. \ref{fig:epoch_elec} in all other cases the MultiResUNet model consistently outperformed the classical U-Net model. In spite of lagging behind the U-Net at the beginning for the electron microscopy images (Fig. \ref{fig:epoch_elec}), eventually the MultiResUNet model converges at a better accuracy than U-Net. Another remarkable observation from the experiments is that except for some minor flactuations, the standard deviation of the performance of the MultiResUNet is much smaller; this indicates the reliability and the robustness of the proposed model.

These results, therefore, suggest that using the proposed MultiResUNet architecture, we are likely to obtain superior results in less number of training epochs as compared to the classical U-Net architecture. 

\subsection{MultiResUNet Delineates Faint Boundaries Better}

\begin{figure}[h]
    \centering
    \includegraphics[width=\textwidth]{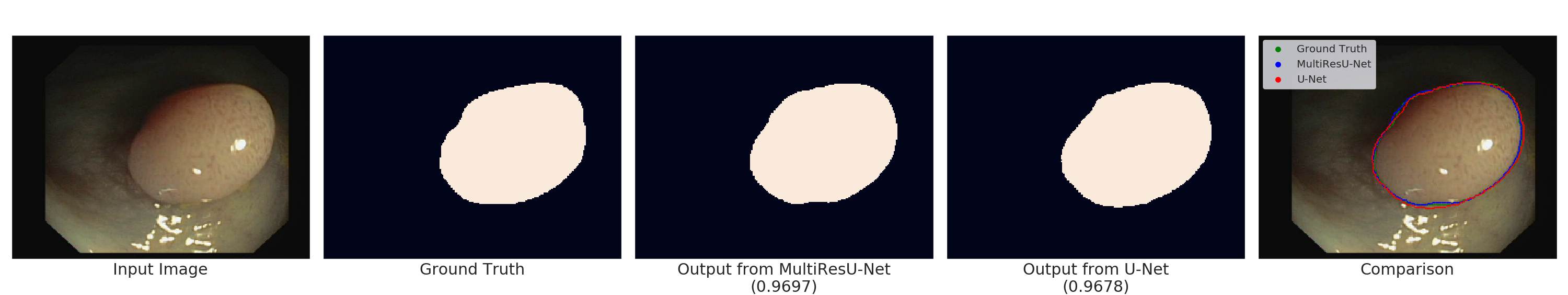}
    \caption{Segmenting a polyp with clearly visiable boundary. U-Net manages to segment the polyp with a high level of performance (J.I. = 0.9678). MultiResUNet segments just slightly better (J.I. = 0.9697)}
    \label{fig:perfect_40}
\end{figure}

Being the current state of the art model for medical image segmentation, U-Net has demonstrated quite satisfactory results in our experiments. For instance, in Fig. \ref{fig:perfect_40}, for a polyp with clearly distinguishable boundary the U-Net model manages to segment it with a high value of Jaccard Index; our proposed model however performs better albeit only slightly.

\begin{figure}[h]
    \centering
    \begin{subfigure}[h]{0.242\textwidth}
        \includegraphics[width=\textwidth]{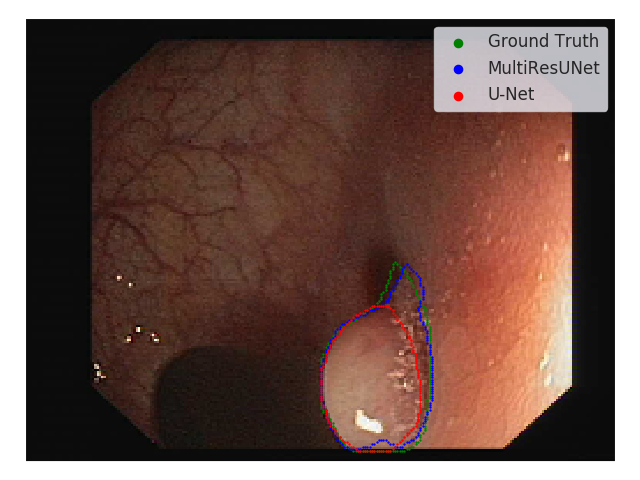}
        \caption{}
        \label{fig:badboundary1}
    \end{subfigure}
    \begin{subfigure}[h]{0.242\textwidth}
        \includegraphics[width=\textwidth]{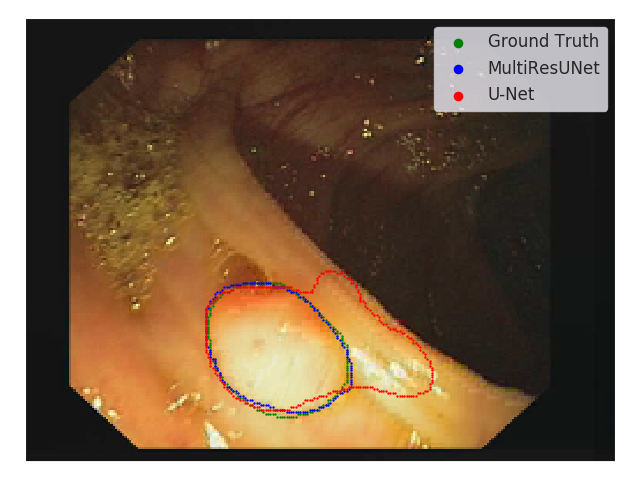}
        \caption{}
        \label{fig:badboundary2}
    \end{subfigure}
    \begin{subfigure}[h]{0.242\textwidth}
        \includegraphics[width=\textwidth]{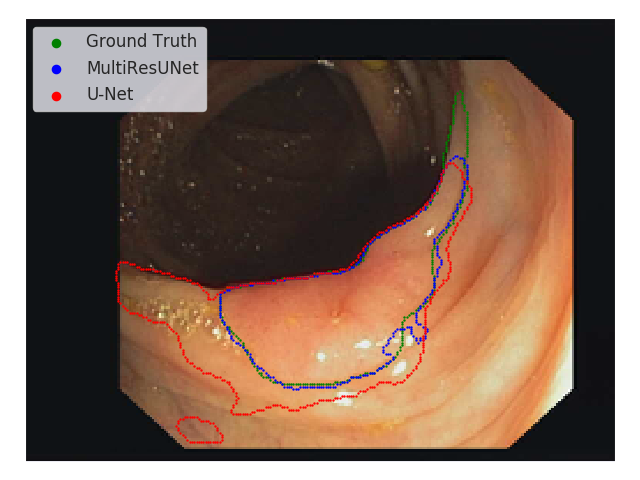}
        \caption{}
        \label{fig:badboundary3}
    \end{subfigure}
    \begin{subfigure}[h]{0.242\textwidth}
        \includegraphics[width=\textwidth]{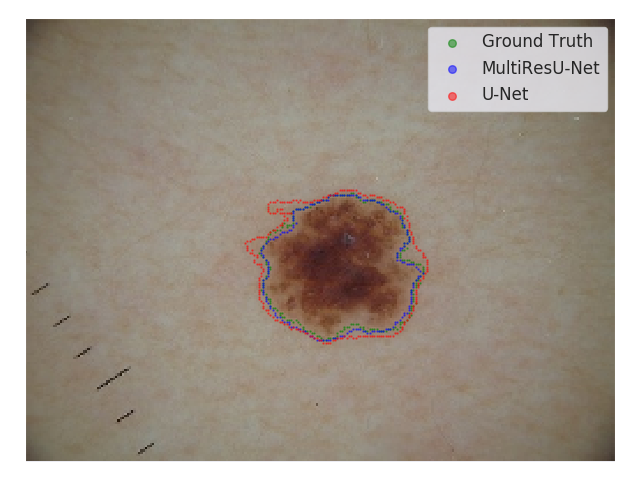}
        \caption{}
        \label{fig:badboundary4}
    \end{subfigure}
    \caption{Segmenting images with vague boundaries. This issue is more prominent for Colon Endoscopy images. U-Net seems to either under-segment (\ref{fig:badboundary1}), or over-segment (\ref{fig:badboundary2}) the polyps. MultiResUNet manages to segment polyps of such situation much better. However, some images are too problematic even for MultiResUNet, but in those cases as well it performs beter than U-Net (\ref{fig:badboundary3}). Even in dermoscopy images, where there exists a clear boundary, U-Net sometimes produced some irregularities along the boundaries, but MultiResUNet was much more robust (\ref{fig:badboundary4}).}
    \label{fig:badboundary}
\end{figure}

But as we study more and more challenging images, especially with not so much conspicuous boundaries, U-Net seems to be struggling a bit (Fig. \ref{fig:badboundary}). The colon polyp images often suffer from the lack of clear boundaries. On such cases, the U-Net model either under-segmented (Fig. \ref{fig:badboundary1}) or over-segmented (Fig. \ref{fig:badboundary2}) the polyps. Our proposed MultiResUNet, on the other hand, performed considerably better in both the cases. However, there are some images where both the models faced complications, but in those cases MultiResUNet's performance was superior(Fig. \ref{fig:badboundary3}). Dermoscopic images have comparatively clearer defined boundaries; still in those cases MultiResUNet delineats the boundaries better (Fig. \ref{fig:badboundary4}). Same was observed for other types of images. We hypothesize that the use of multiple filter sizes allows MultiResUNet to perform better pixel perfect segmentation.

\subsection{MultiResUNet is More Immune to Perturbations}

The core concept of semantic segmentation is to cluster the homologous regions of an image together. However, often in real world medical images the homologous regions get deviated due to various types of noises, artifacts and irregularities in general. This makes it challenging to distinguish between the region of interest and background in medical images. As a result, instead of obtaining a continuous segmented region, we are often left with a collection of fractured segmented regions. At the other extreme, due to textures and perturbations the plain background sometimes appear similar to the region of interest. These two cases lead to loss of information and false classifications respectively. Fortunately, the Dermatoscopy image dataset we have used contains images with such confusing cases, allowing us to analyze and compare the behaviour and performance of the two models.

\begin{figure}[h]
    \centering
    \begin{subfigure}[h]{0.242\textwidth}
        \includegraphics[width=\textwidth]{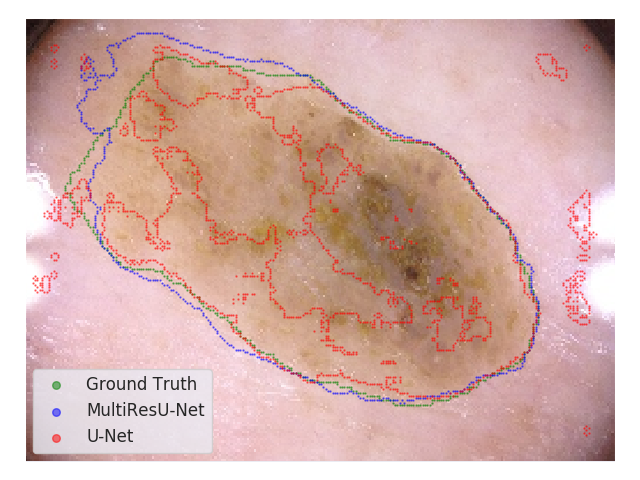}
        \caption{}
        \label{fig:perturbation1}
    \end{subfigure}
    \begin{subfigure}[h]{0.242\textwidth}
        \includegraphics[width=\textwidth]{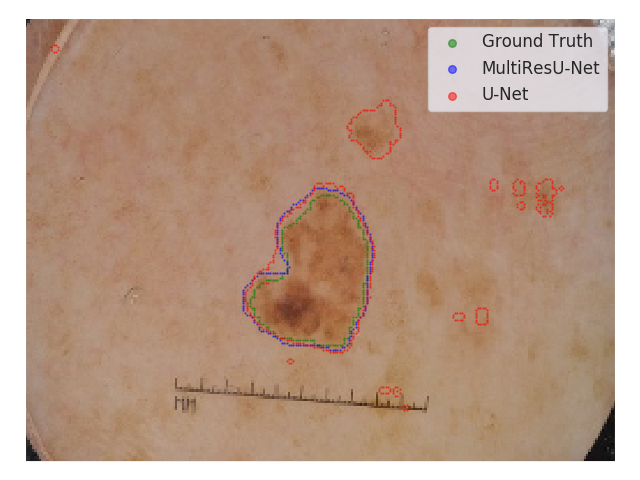}
        \caption{}
        \label{fig:perturbation2}
    \end{subfigure}
    \begin{subfigure}[h]{0.242\textwidth}
        \includegraphics[width=\textwidth]{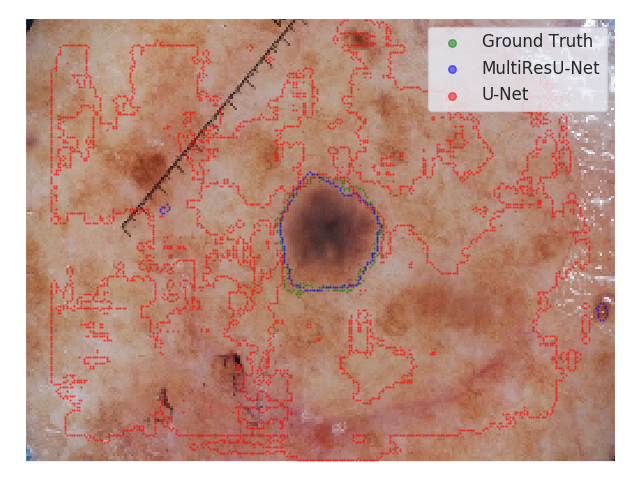}
        \caption{}
        \label{fig:perturbation3}
    \end{subfigure}
    \begin{subfigure}[h]{0.242\textwidth}
        \includegraphics[width=\textwidth]{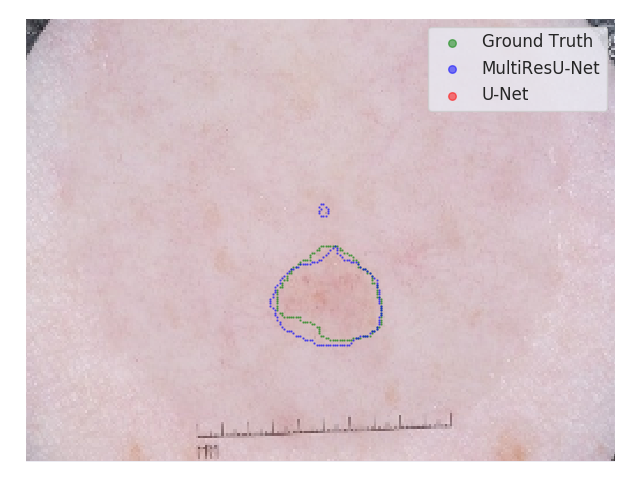}
        \caption{}
        \label{fig:perturbation4}
    \end{subfigure}
    \caption{Segmenting images with irregularities. For images where the foreground is not consistent throughout, instead of segmenting it as a continuous region, U-Net seems to have predicted a set of small regions (\ref{fig:perturbation1}). For images with rough backgrounds U-Net classified them as the foreground (\ref{fig:perturbation2}), the more irregular the background, the more false predictions were made (\ref{fig:perturbation3}). To the other extreme, for images where difference between foreground and background is too subtle, U-Net missed the foreground completely (\ref{fig:perturbation4}). Though the segmentations produced by MultiResUNet in these challenging cases are not perfect, they were consistently better than that of the U-Net.}
    \label{fig:perturbation}
\end{figure}

In spite of segmenting the images with near consistent background and approximately undeviating foreground with almost perfection, the baseline U-Net model seems to struggle quite a bit in the presence of perturbations in images (Fig. \ref{fig:perturbation}). In images where the foreground object tend to vary a bit therein, U-Net, was unable to segment the forground as a continuous region. It rather predicted a set of scattered regions (Fig. \ref{fig:perturbation1}), confusing the forground as background and thus caused the loss of some valuable information. On the other hand, for images where the background is not uniform, the U-Net model seems to make some false predictions (Fig. \ref{fig:perturbation2}). The more rough the background becomes, the more false predictions are made (Fig. \ref{fig:perturbation3}). Furthermore, in some dreadfully adverse situations, where due to irregularity the difference between background and forground are too subtle, the U-Net model failed to make any predictions at all (Fig. \ref{fig:perturbation4}). Although in such challenging cases the segmentation of MultiResUNet is not perfect, it performs far superior than the classical U-Net model as shown in Fig. \ref{fig:perturbation}. It is worth noting here that in the initial stages of our experiments, prior to using the $ResPath$s, our proposed model was also being affected by such perturbations. Therefore, we conjecture that applying additional non-linear operations on the encoder feature maps makes it robust against perturbations.

\subsection{MultiResUNet is More Reliable Against Outliers}

\begin{figure}[h]
    \centering
    \includegraphics[width=\textwidth]{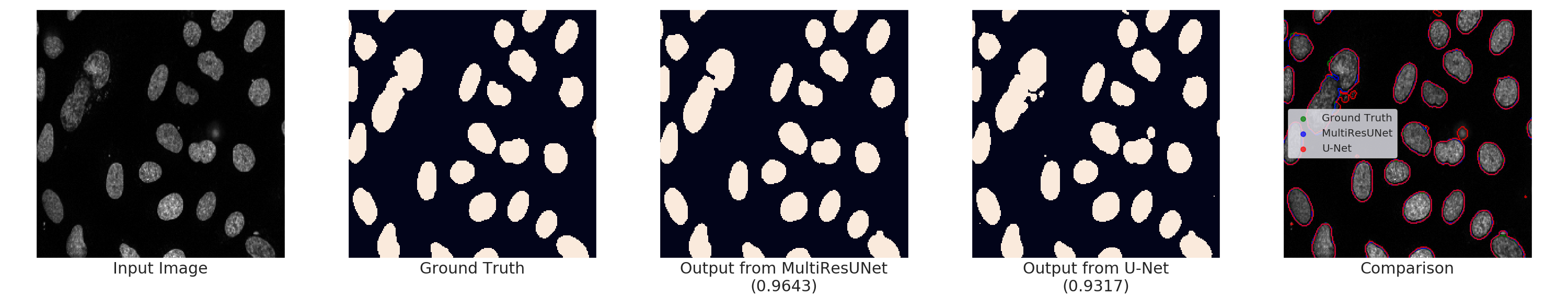}
    \caption{Segmenting images containing outliers. In the fluorescence microscopy images, the exist some bright particles, visually very similar to the cell nuclei under analysis. Although MultiResUNet can identify and reject those outliers, U-Net seems to have missclassified them.}
    \label{fig:outlier}
\end{figure}

Often in medical images some outliers are present, which in spite of being visually quite similar, are different from what we are interested in segmenting. Particularly, in the Fluorescence Microscopy image dataset, there exist some images with bright objects, that are apparently almost indistinguishable from the actual nuclei. Such an example is shown in Fig. \ref{fig:outlier}.

It can be observed that the input image is infected with some small particles that are not actual cell nuclei. However, if we study the segmentation mask generated from U-Net, it turns out that U-Net has mistakenly predicted those outlier particles to be cell nuclei. On the other hand, our proposed MultiResUNet seems to be able to reject those outliers. Since the outliers are pretty tiny, false predictions made by the U-Net model does not hurt the value of Jaccard Index that much ($0.9317$ instead of $0.9643$, when outliers are filtered out). Nevertheless, being able to segregate these outliers are of substantial significance. It can be noted that, similar types of visually similar outliers were present in other datasets as well, and MultiResUNet was able to segment the images reliably without making false predictions.

\subsection{Note on Segmenting the Majority Class}

\begin{figure}[h]
    \centering
    \includegraphics[width=\textwidth]{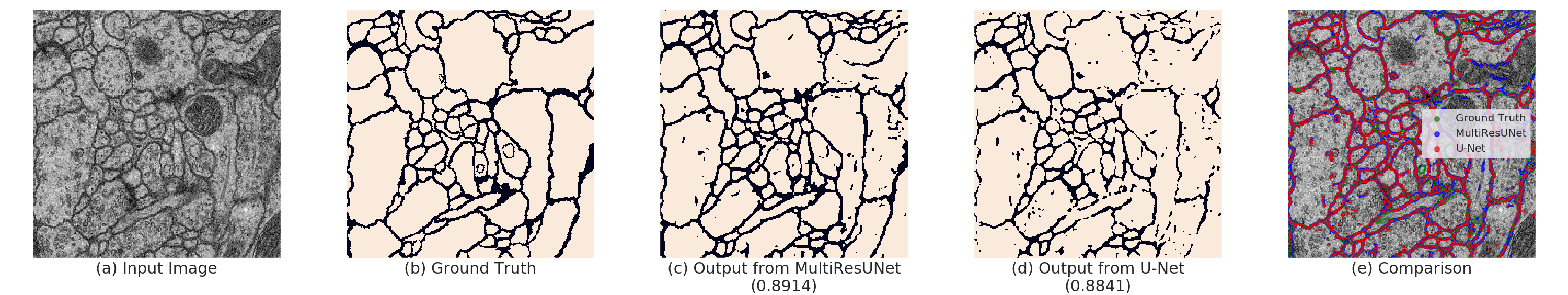}
    \caption{Segmenting the Majority Class. Here we can observe that the region of interest comprises most part of the image. Despite the values of Jacard Index for both U-Net and MultiResUNet are quite similar, visually the segmentation mask are very different. It can be seen that segmentation mask generated from MultiResUNet captures most of the fine separating lines, but U-Net tends to miss them. Moreover, there are some clusters of background pixels which although are missed by U-Net, have been roughly identified by MultiResUNet. Since the class being segmented is the majority class, the values of Jacard Index are inflated.}
    \label{fig:majority}
\end{figure}

The Electron Microscopy dataset we used in our experiments is quite interesting and unorthodox as in this dataset the region of interest under consideration actually comprises the majority of the images. This is a rare incident since usually the region of interest consists of a small portion of the image. This brings a different type of challenge as in such a case the models tend to over-segment the images unnecessarily to minimize the losses during training. A relevant example is presented in Fig. \ref{fig:majority}.

Here, it can be observed quite astonishingly that the majority of the image is actually forground (Fig. \ref{fig:majority}b), with some narrow separations among them by the background, i.e., membranes in this context. If we analyze the segmentation predicted by U-Net, it appears that those fine lines of separation have often been missed (Fig. \ref{fig:majority}d). MultiResUNet, on the other hand, have managed to segment the regions with properly defined separations among them (Fig. \ref{fig:majority}c). Also, it can be observed that there are some small clusters of background pixels, which have been captured with some success in the segmentation mask predicted by MultiResUNet, but are almost non-existent in the segmentation performed by U-Net. Furthermore, the result generated by MultiResUNet seems to be more immune to the noises present in the image. 

Despite the two segmentations (i.e. the results of MultiResUNet and U-Net) are very different from each other, the values of the respective Jaccard Index are quite alike ($0.8914$ and $0.8841$ as shown in Fig. \ref{fig:majority}). This is due to the fact that, the metric Jaccard Index has been inflated with the results of segmenting the majority class of the image. Therefore, the value of Jaccard Index is not that much of a proper representative of accuracy while segmenting the majority class. Despite predicting much inferior segmentations, for this reason the Jaccard Index of U-Net are very close to that of MultiResUNet. Thus, among all the different datasets, the improvement in terms of metric has been underwhelming in this dataset but the predicted segmentations are more accurate visually.

\section{Conclusion}
\label{sec:discussions}

In this work, we started by analyzing the U-Net architecture diligently, with the hope of finding potential rooms for improvement. We anticipated some discrepancy between the features passed from the encoder network and the features propagating through the decoder network. To reconcile these two incompatible sets of features, we proposed \textit{Res} paths, that introduce some additional processing to make the two feature maps more homogeneous. Furthermore, to augment U-Net with the ability of multi-resolutiona analysis, we proposed \textit{MultiRes} blocks. We took inspirations from Inception blocks and formulated a compact analogous structure, that was lightweight and demanded less memory. Incorporating these modifications, we developed a novel architecture, MultiResUNet.

Among the handful publicly available biomedical image datasets, we selected the ones that were drastically different from each other. Additionally each of these datasets poses a separate challenge of its own. The Murphy Lab Fluorescence Microscopy dataset is possibly the simplest dataset for performing segmentation, having an acute difference in contrast between the forground, i.e., the cell nuclei and the background, but contains some outliers. The CVC-ClinicDB dataset contains colon endoscopy images where the boundaries between the polyps and the background are so vague that often it becomes difficult to distinguish even for a trained operator. In addition, the polyps are diverse in terms of shape, size, structure, orientation etc., making this dataset indeed a challenging one. On the other hand, the dermoscopy dataset of ISIC-2018 competition contains images of poor contrast to the extent that sometimes the skin lesions seem identical to the background and vice versa. Moreover, various types of textures present in both the background and the foreground make pattern recognition quite difficult. ISBI-2012 electron microscopy dataset presents a different type of challenge. In this dataset the region being segmented covers the majority of the image; thus a tendency is observed to over-segment the images. The BraTS17 MRI dataset, on the other hand contains multimodal 3D images, which is a different problem alltogether.

For perfect or near perfect images U-Net manages to perform segmentation with remarkable accuracy. Our proposed architecture performs only slightly better than U-Net in those cases. However, for intricate images suffering from noises, perturbations, lack of clear boundaries etc., the gain in performance by MultiResUNet dramatically increases. More specifically, for the five datasets a relative improvement in performance of 10.15\%, 5.07\%, 2.63\%, 1.41\%, and 0.62\% were observed in using MultiResUNet over U-Net (Table \ref{table:comparison}). Not only the segmentations generated by MultiResUNet attain higher score in the evaluation metric, they are also visually more similar to the ground truth. Furthermore, on the very challenging images U-Net tended to over-segment, under-segment, make false predictions and even miss the objects completely. On the contrary, in the experiments MultiResUNet appeared to be more reliable and robust. MultiResUNet managed to detect even the most subtle boundaries, was resilient in segmenting images with a lot of perturbations, and was rejectable to the outliers. Even in segmenting the majority class, where the U-Net tended to over-segment, MultiResUNet managed to capture the fine details. Furthermore, the 3D adaptation of MultiResUNet performed better than 3D U-Net, which is not just a straightforward 3D implementation of the U-Net, but an enhanced and improved version. It should be noted that, the segmentations generated by the proposed MultiResUNet were not perfect, but in most of the cases it outperformed the classical U-Net by a large margin.

Therefore, we believe our proposed MultiResUNet architecture can be the potential successor to the classical U-Net architecture. The future direction of this research has several branches. In this work, we have been motivated to keep the number of parameters of our model comparable to that of the U-Net model. However, in future we wish to conduct experiments to determine the best set of hyperparameters for the model more exhaustively. Moreover, we would like to evaluate our model performance on medical images originating from other modalities as well. Furthermore, we are interested in experimenting by applying several domain and application specific pre-processing and post-processing schemes to our model. We believe fusing our model to a domain specific expert knowledge based pipeline, and coupling it with proper post-processing stages will improve our model performance further, and allow us to develop better segmentation methods for diversified applications.

\section*{Acknowledgements}

The Titan Xp GPU used for this research was the generous donation of NVIDIA Corporation.

\section*{Supplementary information}
\textbf{Supplementary Material 1}: contains the links to the weights and parameters of the best performing models in each fold for all the datasets.\\
\textbf{Supplementary Material 2}: provides the random data splits of the datasets obtained using standard methods of Scikit-learn \cite{scikit-learn}.\\
\textbf{Supplementary Material 3}: presents the detailed experimental results.

\bibliographystyle{unsrt}
\bibliography{main.bib}

\begin{thebibliography}{10}

\bibitem{schindelin2015imagej}
Johannes Schindelin, Curtis~T Rueden, Mark~C Hiner, and Kevin~W Eliceiri.
\newblock The imagej ecosystem: an open platform for biomedical image analysis.
\newblock {\em Molecular reproduction and development}, 82(7-8):518--529, 2015.

\bibitem{mcguinness2010comparative}
Kevin McGuinness and Noel~E O’connor.
\newblock A comparative evaluation of interactive segmentation algorithms.
\newblock {\em Pattern Recognition}, 43(2):434--444, 2010.

\bibitem{codella2018skin}
Noel~CF Codella, David Gutman, M~Emre Celebi, Brian Helba, Michael~A Marchetti,
  Stephen~W Dusza, Aadi Kalloo, Konstantinos Liopyris, Nabin Mishra, Harald
  Kittler, et~al.
\newblock Skin lesion analysis toward melanoma detection: A challenge at the
  2017 international symposium on biomedical imaging (isbi), hosted by the
  international skin imaging collaboration (isic).
\newblock In {\em Biomedical Imaging (ISBI 2018), 2018 IEEE 15th International
  Symposium on}, pages 168--172. IEEE, 2018.

\bibitem{yang2018autosegmentation}
Jinzhong Yang, Harini Veeraraghavan, Samuel~G Armato~III, Keyvan Farahani,
  Justin~S Kirby, Jayashree Kalpathy-Kramer, Wouter van Elmpt, Andre Dekker,
  Xiao Han, Xue Feng, et~al.
\newblock Autosegmentation for thoracic radiation treatment planning: A grand
  challenge at aapm 2017.
\newblock {\em Medical physics}, 2018.

\bibitem{naik2008automated}
Shivang Naik, Scott Doyle, Shannon Agner, Anant Madabhushi, Michael Feldman,
  and John Tomaszewski.
\newblock Automated gland and nuclei segmentation for grading of prostate and
  breast cancer histopathology.
\newblock In {\em Biomedical Imaging: From Nano to Macro, 2008. ISBI 2008. 5th
  IEEE International Symposium on}, pages 284--287. IEEE, 2008.

\bibitem{rouhi2015benign}
Rahimeh Rouhi, Mehdi Jafari, Shohreh Kasaei, and Peiman Keshavarzian.
\newblock Benign and malignant breast tumors classification based on region
  growing and cnn segmentation.
\newblock {\em Expert Systems with Applications}, 42(3):990--1002, 2015.

\bibitem{pham2000current}
Dzung~L Pham, Chenyang Xu, and Jerry~L Prince.
\newblock Current methods in medical image segmentation.
\newblock {\em Annual review of biomedical engineering}, 2(1):315--337, 2000.

\bibitem{mesejo2015biomedical}
Pablo Mesejo, Andrea Valsecchi, Linda Marrakchi-Kacem, Stefano Cagnoni, and
  Sergio Damas.
\newblock Biomedical image segmentation using geometric deformable models and
  metaheuristics.
\newblock {\em Computerized Medical Imaging and Graphics}, 43:167--178, 2015.

\bibitem{zheng2015image}
Yuhui Zheng, Byeungwoo Jeon, Danhua Xu, QM~Wu, and Hui Zhang.
\newblock Image segmentation by generalized hierarchical fuzzy c-means
  algorithm.
\newblock {\em Journal of Intelligent \& Fuzzy Systems}, 28(2):961--973, 2015.

\bibitem{lecun2015deep}
Yann LeCun, Yoshua Bengio, and Geoffrey Hinton.
\newblock Deep learning.
\newblock {\em nature}, 521(7553):436, 2015.

\bibitem{lecun1998gradient}
Yann LeCun, L{\'e}on Bottou, Yoshua Bengio, and Patrick Haffner.
\newblock Gradient-based learning applied to document recognition.
\newblock {\em Proceedings of the IEEE}, 86(11):2278--2324, 1998.

\bibitem{krizhevsky2012imagenet}
Alex Krizhevsky, Ilya Sutskever, and Geoffrey~E Hinton.
\newblock Imagenet classification with deep convolutional neural networks.
\newblock In {\em Advances in neural information processing systems}, pages
  1097--1105, 2012.

\bibitem{sermanet2013overfeat}
Pierre Sermanet, David Eigen, Xiang Zhang, Micha{\"e}l Mathieu, Rob Fergus, and
  Yann LeCun.
\newblock Overfeat: Integrated recognition, localization and detection using
  convolutional networks.
\newblock {\em arXiv preprint arXiv:1312.6229}, 2013.

\bibitem{simonyan2014very}
Karen Simonyan and Andrew Zisserman.
\newblock Very deep convolutional networks for large-scale image recognition.
\newblock {\em arXiv preprint arXiv:1409.1556}, 2014.

\bibitem{szegedy2015going}
Christian Szegedy, Wei Liu, Yangqing Jia, Pierre Sermanet, Scott Reed, Dragomir
  Anguelov, Dumitru Erhan, Vincent Vanhoucke, and Andrew Rabinovich.
\newblock Going deeper with convolutions.
\newblock In {\em Proceedings of the IEEE conference on computer vision and
  pattern recognition}, pages 1--9, 2015.

\bibitem{he2016deep}
Kaiming He, Xiangyu Zhang, Shaoqing Ren, and Jian Sun.
\newblock Deep residual learning for image recognition.
\newblock In {\em Proceedings of the IEEE conference on computer vision and
  pattern recognition}, pages 770--778, 2016.

\bibitem{szegedy2017inception}
Christian Szegedy, Sergey Ioffe, Vincent Vanhoucke, and Alexander~A Alemi.
\newblock Inception-v4, inception-resnet and the impact of residual connections
  on learning.
\newblock In {\em AAAI}, volume~4, page~12, 2017.

\bibitem{ciresan2012deep}
Dan Ciresan, Alessandro Giusti, Luca~M Gambardella, and J{\"u}rgen Schmidhuber.
\newblock Deep neural networks segment neuronal membranes in electron
  microscopy images.
\newblock In {\em Advances in neural information processing systems}, pages
  2843--2851, 2012.

\bibitem{long2015fully}
Jonathan Long, Evan Shelhamer, and Trevor Darrell.
\newblock Fully convolutional networks for semantic segmentation.
\newblock In {\em Proceedings of the IEEE conference on computer vision and
  pattern recognition}, pages 3431--3440, 2015.

\bibitem{badrinarayanan2015segnet}
Vijay Badrinarayanan, Alex Kendall, and Roberto Cipolla.
\newblock Segnet: A deep convolutional encoder-decoder architecture for image
  segmentation.
\newblock {\em arXiv preprint arXiv:1511.00561}, 2015.

\bibitem{chen2018deeplab}
Liang-Chieh Chen, George Papandreou, Iasonas Kokkinos, Kevin Murphy, and Alan~L
  Yuille.
\newblock Deeplab: Semantic image segmentation with deep convolutional nets,
  atrous convolution, and fully connected crfs.
\newblock {\em IEEE transactions on pattern analysis and machine intelligence},
  40(4):834--848, 2018.

\bibitem{litjens2017survey}
Geert Litjens, Thijs Kooi, Babak~Ehteshami Bejnordi, Arnaud Arindra~Adiyoso
  Setio, Francesco Ciompi, Mohsen Ghafoorian, Jeroen~Awm Van Der~Laak, Bram
  Van~Ginneken, and Clara~I S{\'a}nchez.
\newblock A survey on deep learning in medical image analysis.
\newblock {\em Medical image analysis}, 42:60--88, 2017.

\bibitem{anwar2018medical}
Syed~Muhammad Anwar, Muhammad Majid, Adnan Qayyum, Muhammad Awais, Majdi
  Alnowami, and Muhammad~Khurram Khan.
\newblock Medical image analysis using convolutional neural networks: a review.
\newblock {\em Journal of medical systems}, 42(11):226, 2018.

\bibitem{ronneberger2015u}
Olaf Ronneberger, Philipp Fischer, and Thomas Brox.
\newblock U-net: Convolutional networks for biomedical image segmentation.
\newblock In {\em International Conference on Medical image computing and
  computer-assisted intervention}, pages 234--241. Springer, 2015.

\bibitem{christ2016automatic}
Patrick~Ferdinand Christ, Mohamed Ezzeldin~A Elshaer, Florian Ettlinger, Sunil
  Tatavarty, Marc Bickel, Patrick Bilic, Markus Rempfler, Marco Armbruster,
  Felix Hofmann, Melvin D’Anastasi, et~al.
\newblock Automatic liver and lesion segmentation in ct using cascaded fully
  convolutional neural networks and 3d conditional random fields.
\newblock In {\em International Conference on Medical Image Computing and
  Computer-Assisted Intervention}, pages 415--423. Springer, 2016.

\bibitem{lin2017skin}
Bill~S Lin, Kevin Michael, Shivam Kalra, and Hamid~R Tizhoosh.
\newblock Skin lesion segmentation: U-nets versus clustering.
\newblock In {\em 2017 IEEE Symposium Series on Computational Intelligence
  (SSCI)}, pages 1--7. IEEE, 2017.

\bibitem{sirinukunwattana2017gland}
Korsuk Sirinukunwattana, Josien~PW Pluim, Hao Chen, Xiaojuan Qi, Pheng-Ann
  Heng, Yun~Bo Guo, Li~Yang Wang, Bogdan~J Matuszewski, Elia Bruni, Urko
  Sanchez, et~al.
\newblock Gland segmentation in colon histology images: The glas challenge
  contest.
\newblock {\em Medical image analysis}, 35:489--502, 2017.

\bibitem{cciccek20163d}
{\"O}zg{\"u}n {\c{C}}i{\c{c}}ek, Ahmed Abdulkadir, Soeren~S Lienkamp, Thomas
  Brox, and Olaf Ronneberger.
\newblock 3d u-net: learning dense volumetric segmentation from sparse
  annotation.
\newblock In {\em International Conference on Medical Image Computing and
  Computer-Assisted Intervention}, pages 424--432. Springer, 2016.

\bibitem{merkow2016dense}
Jameson Merkow, Alison Marsden, David Kriegman, and Zhuowen Tu.
\newblock Dense volume-to-volume vascular boundary detection.
\newblock In {\em International Conference on Medical Image Computing and
  Computer-Assisted Intervention}, pages 371--379. Springer, 2016.

\bibitem{setio2017validation}
Arnaud Arindra~Adiyoso Setio, Alberto Traverso, Thomas De~Bel, Moira~SN Berens,
  Cas van~den Bogaard, Piergiorgio Cerello, Hao Chen, Qi~Dou, Maria~Evelina
  Fantacci, Bram Geurts, et~al.
\newblock Validation, comparison, and combination of algorithms for automatic
  detection of pulmonary nodules in computed tomography images: the luna16
  challenge.
\newblock {\em Medical image analysis}, 42:1--13, 2017.

\bibitem{yu2017volumetric}
Lequan Yu, Xin Yang, Hao Chen, Jing Qin, and Pheng-Ann Heng.
\newblock Volumetric convnets with mixed residual connections for automated
  prostate segmentation from 3d mr images.
\newblock In {\em AAAI}, pages 66--72, 2017.

\bibitem{zeiler2010deconvolutional}
M.~D. Zeiler, D.~Krishnan, G.~W. Taylor, and R.~Fergus.
\newblock Deconvolutional networks.
\newblock In {\em 2010 IEEE Computer Society Conference on Computer Vision and
  Pattern Recognition}, pages 2528--2535, June 2010.

\bibitem{drozdzal2016importance}
Michal Drozdzal, Eugene Vorontsov, Gabriel Chartrand, Samuel Kadoury, and Chris
  Pal.
\newblock The importance of skip connections in biomedical image segmentation.
\newblock In {\em Deep Learning and Data Labeling for Medical Applications},
  pages 179--187. Springer, 2016.

\bibitem{szegedy2016rethinking}
Christian Szegedy, Vincent Vanhoucke, Sergey Ioffe, Jon Shlens, and Zbigniew
  Wojna.
\newblock Rethinking the inception architecture for computer vision.
\newblock In {\em Proceedings of the IEEE conference on computer vision and
  pattern recognition}, pages 2818--2826, 2016.

\bibitem{ioffe2015batch}
Sergey Ioffe and Christian Szegedy.
\newblock Batch normalization: Accelerating deep network training by reducing
  internal covariate shift.
\newblock {\em arXiv preprint arXiv:1502.03167}, 2015.

\bibitem{coelho2009nuclear}
Lu{\'\i}s~Pedro Coelho, Aabid Shariff, and Robert~F Murphy.
\newblock Nuclear segmentation in microscope cell images: a hand-segmented
  dataset and comparison of algorithms.
\newblock In {\em Biomedical Imaging: From Nano to Macro, 2009. ISBI'09. IEEE
  International Symposium on}, pages 518--521. IEEE, 2009.

\bibitem{serre2007robust}
Thomas Serre, Lior Wolf, Stanley Bileschi, Maximilian Riesenhuber, and Tomaso
  Poggio.
\newblock Robust object recognition with cortex-like mechanisms.
\newblock {\em IEEE Transactions on Pattern Analysis \& Machine Intelligence},
  (3):411--426, 2007.

\bibitem{wang2018understanding}
Panqu Wang, Pengfei Chen, Ye~Yuan, Ding Liu, Zehua Huang, Xiaodi Hou, and
  Garrison Cottrell.
\newblock Understanding convolution for semantic segmentation.
\newblock In {\em 2018 IEEE Winter Conference on Applications of Computer
  Vision (WACV)}, pages 1451--1460. IEEE, 2018.

\bibitem{mao2016image}
Xiao-Jiao Mao, Chunhua Shen, and Yu-Bin Yang.
\newblock Image restoration using convolutional auto-encoders with symmetric
  skip connections.
\newblock {\em arXiv preprint arXiv:1606.08921}, 2016.

\bibitem{arganda2015crowdsourcing}
Ignacio Arganda-Carreras, Srinivas~C Turaga, Daniel~R Berger, Dan
  Cire{\c{s}}an, Alessandro Giusti, Luca~M Gambardella, J{\"u}rgen Schmidhuber,
  Dmitry Laptev, Sarvesh Dwivedi, Joachim~M Buhmann, et~al.
\newblock Crowdsourcing the creation of image segmentation algorithms for
  connectomics.
\newblock {\em Frontiers in neuroanatomy}, 9:142, 2015.

\bibitem{cardona2010integrated}
Albert Cardona, Stephan Saalfeld, Stephan Preibisch, Benjamin Schmid, Anchi
  Cheng, Jim Pulokas, Pavel Tomancak, and Volker Hartenstein.
\newblock An integrated micro-and macroarchitectural analysis of the drosophila
  brain by computer-assisted serial section electron microscopy.
\newblock {\em PLoS biology}, 8(10):e1000502, 2010.

\bibitem{tschandl2018ham10000}
Philipp Tschandl, Cliff Rosendahl, and Harald Kittler.
\newblock The ham10000 dataset: A large collection of multi-source
  dermatoscopic images of common pigmented skin lesions.
\newblock {\em arXiv preprint arXiv:1803.10417}, 2018.

\bibitem{bernal2015wm}
Jorge Bernal, F~Javier S{\'a}nchez, Gloria Fern{\'a}ndez-Esparrach, Debora Gil,
  Cristina Rodr{\'\i}guez, and Fernando Vilari{\~n}o.
\newblock Wm-dova maps for accurate polyp highlighting in colonoscopy:
  Validation vs. saliency maps from physicians.
\newblock {\em Computerized Medical Imaging and Graphics}, 43:99--111, 2015.

\bibitem{menze2015multimodal}
Bjoern~H Menze, Andras Jakab, Stefan Bauer, Jayashree Kalpathy-Cramer, Keyvan
  Farahani, Justin Kirby, Yuliya Burren, Nicole Porz, Johannes Slotboom, Roland
  Wiest, et~al.
\newblock The multimodal brain tumor image segmentation benchmark (brats).
\newblock {\em IEEE transactions on medical imaging}, 34(10):1993, 2015.

\bibitem{bakas2017advancing}
Spyridon Bakas, Hamed Akbari, Aristeidis Sotiras, Michel Bilello, Martin
  Rozycki, Justin~S Kirby, John~B Freymann, Keyvan Farahani, and Christos
  Davatzikos.
\newblock Advancing the cancer genome atlas glioma mri collections with expert
  segmentation labels and radiomic features.
\newblock {\em Scientific data}, 4:170117, 2017.

\bibitem{van2007python}
Guido Van~Rossum et~al.
\newblock Python programming language.
\newblock In {\em USENIX Annual Technical Conference}, volume~41, page~36,
  2007.

\bibitem{chollet2015keras}
Fran{\c{c}}ois Chollet et~al.
\newblock Keras, 2015.

\bibitem{abadi2016tensorflow}
Mart{\'\i}n Abadi, Paul Barham, Jianmin Chen, Zhifeng Chen, Andy Davis, Jeffrey
  Dean, Matthieu Devin, Sanjay Ghemawat, Geoffrey Irving, Michael Isard, et~al.
\newblock Tensorflow: a system for large-scale machine learning.
\newblock In {\em OSDI}, volume~16, pages 265--283, 2016.

\bibitem{kingma2014adam}
Diederik~P Kingma and Jimmy Ba.
\newblock Adam: A method for stochastic optimization.
\newblock {\em arXiv preprint arXiv:1412.6980}, 2014.

\bibitem{duchi2011adaptive}
John Duchi, Elad Hazan, and Yoram Singer.
\newblock Adaptive subgradient methods for online learning and stochastic
  optimization.
\newblock {\em Journal of Machine Learning Research}, 12(Jul):2121--2159, 2011.

\bibitem{tieleman2012lecture}
Tijmen Tieleman and Geoffrey Hinton.
\newblock Lecture 6.5-rmsprop: Divide the gradient by a running average of its
  recent magnitude.
\newblock {\em COURSERA: Neural networks for machine learning}, 4(2):26--31,
  2012.

\bibitem{ruder2016overview}
Sebastian Ruder.
\newblock An overview of gradient descent optimization algorithms.
\newblock {\em arXiv preprint arXiv:1609.04747}, 2016.

\bibitem{kohavi1995study}
Ron Kohavi et~al.
\newblock A study of cross-validation and bootstrap for accuracy estimation and
  model selection.
\newblock In {\em Ijcai}, volume~14, pages 1137--1145. Montreal, Canada, 1995.

\bibitem{scikit-learn}
F.~Pedregosa, G.~Varoquaux, A.~Gramfort, V.~Michel, B.~Thirion, O.~Grisel,
  M.~Blondel, P.~Prettenhofer, R.~Weiss, V.~Dubourg, J.~Vanderplas, A.~Passos,
  D.~Cournapeau, M.~Brucher, M.~Perrot, and E.~Duchesnay.
\newblock Scikit-learn: Machine learning in {P}ython.
\newblock {\em Journal of Machine Learning Research}, 12:2825--2830, 2011.

\end{thebibliography}

\end{document}